\documentclass{article}
\usepackage{iclr2027_conference,times}

\usepackage{amsmath,amsfonts,bm}

\def\eqref#1{equation~\ref{#1}}

\def\1{\bm{1}}

\DeclareMathAlphabet{\mathsfit}{\encodingdefault}{\sfdefault}{m}{sl}
\SetMathAlphabet{\mathsfit}{bold}{\encodingdefault}{\sfdefault}{bx}{n}

\usepackage{amsmath}
\usepackage{amssymb}
\usepackage{amsthm}
\usepackage{algorithm}
\usepackage{algpseudocode}

\usepackage{newtxmath}

\usepackage{booktabs}
\usepackage{array}
\usepackage{tabularx}
\usepackage{multirow}
\usepackage{graphicx}
\usepackage{xcolor}
\usepackage{colortbl}
\usepackage{enumitem}
\usepackage{microtype}
\usepackage{float}
\usepackage{adjustbox}
\usepackage{wrapfig}
\usepackage{capt-of}
\usepackage[font=small]{caption}
\usepackage{needspace}
\makeatletter
\newcommand{\RGDTfinishwrap}{\par
  \ifnum\c@WF@wrappedlines>\@ne
    \begingroup
      \@tempdima\baselineskip
      \multiply\@tempdima\c@WF@wrappedlines
      \advance\@tempdima-\baselineskip
      \vskip\@tempdima
    \endgroup
  \fi
  \WFclear}
\makeatother
\usepackage{placeins}
\usepackage{hyperref}
\usepackage{url}
\usepackage[most]{tcolorbox}
\usepackage{fontawesome5}
\hypersetup{hidelinks}
\graphicspath{{figures/}}

\AddToHook{env/table/begin}{\setlength{\abovecaptionskip}{0pt}\setlength{\belowcaptionskip}{4pt}}
\AddToHook{env/table*/begin}{\setlength{\abovecaptionskip}{0pt}\setlength{\belowcaptionskip}{4pt}}
\AddToHook{env/wraptable/begin}{\setlength{\abovecaptionskip}{0pt}\setlength{\belowcaptionskip}{4pt}}

\definecolor{rgdtnavy}{HTML}{172B3A}
\definecolor{rgdtink}{HTML}{2B3942}
\definecolor{rgdtmuted}{HTML}{6D7880}
\definecolor{rgdtline}{HTML}{CDD4D8}
\definecolor{rgdtblue}{HTML}{2E6F86}
\definecolor{rgdtteal}{HTML}{28766F}
\definecolor{rgdtgreen}{HTML}{3D7E62}
\definecolor{rgdtamber}{HTML}{B77A2A}
\definecolor{rgdtred}{HTML}{A94B48}
\definecolor{rgdtlight}{HTML}{F4F6F7}
\definecolor{rgdttablehead}{HTML}{54467F}
\definecolor{rgdttableheadbg}{HTML}{F2F0F8}
\definecolor{rgdttablegroup}{HTML}{54467F}
\definecolor{rgdttablegroupbg}{HTML}{ECEAF5}
\definecolor{rgdttabletotalbg}{HTML}{F7F6FA}
\definecolor{rgdtpositive}{HTML}{285C45}
\definecolor{rgdtpositivebg}{HTML}{DDEDE3}
\definecolor{rgdtpositivebgpale}{HTML}{EFF6F1}
\definecolor{rgdtnegative}{HTML}{8E3F48}
\definecolor{rgdtnegativebg}{HTML}{F3DEE1}
\definecolor{rgdtnegativebgpale}{HTML}{FAEFF0}
\definecolor{rgdtneutralbadge}{HTML}{4E5963}
\definecolor{rgdtneutralbadgebg}{HTML}{EEF0F4}

\colorlet{rgdtstructure}{rgdtnavy}
\colorlet{rgdtvisible}{rgdtblue}
\colorlet{rgdtdiagnostic}{rgdtteal}
\colorlet{rgdtoffline}{rgdtamber}
\colorlet{rgdtvalid}{rgdtgreen}
\colorlet{rgdtrisk}{rgdtred}

\title{RGDT-Bench: Benchmarking LLM Reasoning\\for Rule-Governed Decisions\\and Their Justifications}

\author{\hspace*{-\tabcolsep}%
\parbox{\dimexpr\textwidth-2\tabcolsep\relax}{\raggedright
Jianpeng Zhao\textsuperscript{1,2,\textdagger}, Haihua Xu\textsuperscript{1}, Haoyang Zhang\textsuperscript{1}, Shuang Qian\textsuperscript{2}, Yixiang Tang\textsuperscript{1}\\
Xintao Wang\textsuperscript{1}, Kun Sun\textsuperscript{2}, Pei Wu\textsuperscript{2}, Shuhan Zhong\textsuperscript{2}, Pengyang Wang\textsuperscript{1,*}\\[5pt]
\normalfont\small
\textsuperscript{1}State Key Laboratory of Internet of Things for Smart City and\\
\hphantom{\textsuperscript{1}}Institute of Smart City Technologies, University of Macau, Macau SAR, China\\[2pt]
\textsuperscript{2}ByteDance, Beijing, China
}
}

\DeclareRobustCommand{\rgdt}{\texorpdfstring{\raisebox{-0.02em}{\includegraphics[height=0.80em]{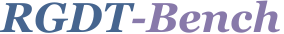}}}{RGDT-Bench}}

\newlength{\rgdttableedgepad}
\newcommand{\tableheaderrow}{}
\newcommand{\tabletotalrow}{\rowcolor{rgdttabletotalbg}}
\newcommand{\tablehead}[1]{{\bfseries\boldmath #1}}
\newcommand{\tablebest}[1]{\textbf{#1}}
\newcommand{\tablesecond}[1]{\underline{#1}}

\newcommand{\sd}[1]{\ensuremath{{}_{\pm #1}}}
\newcommand{\tablegroup}[2]{%
  \rowcolor{rgdttablegroupbg}%
  \multicolumn{#1}{c}{\strut\textbf{\textit{\textcolor{rgdttablegroup}{#2}}}}\\%
}
\newcommand{\rgdtlogodeepseek}{\rule[-0.22em]{0pt}{1.15em}\makebox[1.12em][c]{\raisebox{-0.17em}{\includegraphics[width=0.95em,height=0.95em,keepaspectratio]{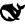}}}\hspace{0.25em}}
\newcommand{\rgdtlogodormscore}{\rule[-0.22em]{0pt}{1.15em}\makebox[1.12em][c]{\raisebox{-0.17em}{\includegraphics[width=0.95em,height=0.95em,keepaspectratio]{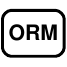}}}\hspace{0.25em}}
\newcommand{\rgdtlogogoogle}{\rule[-0.22em]{0pt}{1.15em}\makebox[1.12em][c]{\raisebox{-0.17em}{\includegraphics[width=0.95em,height=0.95em,keepaspectratio]{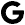}}}\hspace{0.25em}}
\newcommand{\rgdtlogometa}{\rule[-0.22em]{0pt}{1.15em}\makebox[1.12em][c]{\raisebox{-0.17em}{\includegraphics[width=0.95em,height=0.95em,keepaspectratio]{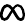}}}\hspace{0.25em}}
\newcommand{\rgdtlogoopenai}{\rule[-0.22em]{0pt}{1.15em}\makebox[1.12em][c]{\raisebox{-0.17em}{\includegraphics[width=0.95em,height=0.95em,keepaspectratio]{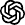}}}\hspace{0.25em}}
\newcommand{\rgdtlogoqwen}{\rule[-0.22em]{0pt}{1.15em}\makebox[1.12em][c]{\raisebox{-0.17em}{\includegraphics[width=0.95em,height=0.95em,keepaspectratio]{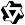}}}\hspace{0.25em}}
\newcommand{\rgdtlogoskywork}{\rule[-0.22em]{0pt}{1.15em}\makebox[1.12em][c]{\raisebox{-0.17em}{\includegraphics[width=0.95em,height=0.95em,keepaspectratio]{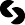}}}\hspace{0.25em}}
\newcommand{\rgdtlogotrmtrace}{\rule[-0.22em]{0pt}{1.15em}\makebox[1.12em][c]{\raisebox{-0.17em}{\includegraphics[width=0.95em,height=0.95em,keepaspectratio]{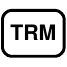}}}\hspace{0.25em}}
\newcommand{\rgdtlogozhipu}{\rule[-0.22em]{0pt}{1.15em}\makebox[1.12em][c]{\raisebox{-0.17em}{\includegraphics[width=0.95em,height=0.95em,keepaspectratio]{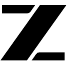}}}\hspace{0.25em}}

\newcommand{\cmark}{\textcolor{rgdtvalid}{\(\checkmark\)}}
\newcommand{\xmark}{\textcolor{rgdtrisk}{\(\times\)}}

\newcommand{\partlabel}[1]{\textbf{\textcolor{rgdtstructure}{(#1)}}}
\DeclareRobustCommand{\tasktrack}[1]{\textit{#1}}

\iclrfinalcopy
\hypersetup{pdftitle={RGDT-Bench: Benchmarking LLM Reasoning for Rule-Governed Decisions and Their Justifications}, pdfauthor={Jianpeng Zhao, Haihua Xu, Haoyang Zhang, Shuang Qian, Yixiang Tang, Xintao Wang, Kun Sun, Pei Wu, Shuhan Zhong, Pengyang Wang}}
\begin{document}
\raggedbottom

\maketitle
\lhead{Preprint}
\begingroup
\renewcommand{\thefootnote}{\fnsymbol{footnote}}
\footnotetext[1]{Corresponding author.}
\footnotetext[2]{Contribution during internship at ByteDance.}
\endgroup

\begin{abstract}
We study reasoning in \textbf{Rule-Governed Decision Tasks (RGDTs)}, where models apply external rules to case facts and justify decisions, as required in policy, contract, and compliance settings.
Beyond the deductive capability emphasized by standard mathematical and logical reasoning tasks, RGDTs require interpreting rules and their applicability, assessing conditions from evidence, combining judgments under rules and exceptions, and providing checkable justifications.
These demands motivate a new benchmark that assesses existing LLMs' reasoning abilities by evaluating both their decisions and their stated grounds.
We introduce \rgdt{}, providing 202.1K condition-level supervision slots across four task tracks and eight supported task--probe combinations that vary access to supporting information.
Label-blind extraction and deterministic checks produce labels for \emph{warrant completeness}: source-referenced coverage and consistency of stated decision grounds.
The benchmark attributes failures to four process layers: rule use, condition, evidence, and aggregation, and checks the final outcome.
Among evaluable correct responses, warrant incompleteness averages 40.2\% across six evaluated LLMs and supported task--probe combinations.
Such warrant incompleteness poses potential safety risks and remains difficult to detect: the best of seventeen existing evaluators reaches only 57.69\% (random: 50\%) task-averaged area under the receiver operating characteristic curve (AUROC).
To address this difficulty, we train a simple reward model with warrant supervision. It achieves 69.24\% task-averaged AUROC among correct answers, exceeding the matched outcome-supervised baseline by 10.37 pp (percentage points) and the best existing evaluator by 11.55 pp. Beyond completeness assessment, the model outperforms both outcome-supervised baselines across nearly all response-selection comparisons, supporting \rgdt{}'s warrant supervision for RGDT reasoning.
\end{abstract}
\vspace{20pt}

\setlength{\parskip}{4pt plus 3pt minus 0.5pt}

\section{Introduction}

From benefit applications and contract review to transaction approval and compliance assessment, many routine decisions depend on how written rules apply to a particular case.
These are \textbf{Rule-Governed Decision Tasks (RGDTs)}: tasks that require both a decision and an explanation of how the governing rules and case facts support it.
Compared with mathematical and formal-logical benchmarks centered on deduction, this setting emphasizes interpreting external decision criteria and establishing how they apply to a case.
It requires four abilities:
\textbf{(1) Rule interpretation and applicability.} Models must identify what the supplied rules require and when they apply.
\textbf{(2) Evidence-grounded condition assessment.} Models must link facts to conditions and judge whether this evidence supports, contradicts, or leaves them unresolved.
\textbf{(3) Decision aggregation.} Models must combine condition judgments according to the governing decision logic, including applicable exceptions.
\textbf{(4) Checkable justification.} Models must explicitly state the required judgments, their supporting rules and evidence, and how they justify the decision.
These demands arise because policy and contract clauses must be linked to differently worded or incomplete case facts.

\Needspace*{250pt}
\begin{wrapfigure}[21]{r}{0.50\textwidth}
  \centering
  \includegraphics[width=\linewidth]{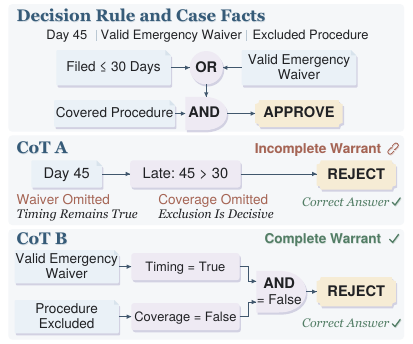}
  \caption{Answer correctness versus warrant completeness in RGDTs.
  Both traces reject correctly, but only CoT B states the waiver and coverage judgments.
  Red notes indicate omissions.}
  \label{fig:teaser}
\end{wrapfigure}

However, final-answer accuracy can conceal failures in these abilities.
In Figure~\ref{fig:teaser}, both traces correctly reject a claim.
One relies on a missed deadline despite a valid waiver, whereas the other recognizes the waiver and bases rejection on the lack of coverage.
These omissions pose potential safety risks, motivating checks of how rules, evidence, and exceptions justify decisions.
These demands motivate a benchmark that assesses LLMs' ability to apply rules to cases, beyond general reasoning scores or final-answer accuracy.
We call the expressed support linking rules and evidence to a decision a \emph{warrant} \citep{toulmin2003uses}.

We introduce \rgdt{}, providing 202.1K condition-level supervision slots across four task tracks: \tasktrack{Policy} \citep{sun2022conditionalqa}, \tasktrack{Contract} \citep{koreeda2021contractnli}, \tasktrack{Regulation} \citep{saeidi2018interpretation}, and \tasktrack{Transaction} \citep{zhou2025rulearena}.
These slots correspond to condition checks within 86,784 model responses (Table~\ref{tab:app_dataset_scale}).
Three warrant-access probes provide evidence and rules matched to the assessed conditions, add selected hard negatives, or retain longer natural document context, yielding eight supported task--probe combinations while preserving reference decisions.
Label-blind claim extraction and deterministic reference checks evaluate these task requirements in the model's stated reasoning, producing labels for \emph{warrant completeness} that we assess through human review.
Process-level failure attribution covers four layers: ``\emph{rule use}'', ``\emph{condition}'', ``\emph{evidence}'', and ``\emph{aggregation}'', alongside a final \emph{outcome} check (Figure~\ref{fig:diagnostic_pipeline}).
Rule-use checks assess the rule support for individual condition judgments; condition and evidence checks assess the judgments and their evidential support; aggregation checks assess whether all required decision conditions are covered and correctly combined.
Together, these checks measure the coverage and consistency of the stated warrant against source-derived references, including applicable exceptions.
Completeness concerns the required grounds, not merely whether the reasoning suffices to reach the answer.

Across six evaluated LLMs, 24.8\%--76.7\% of evaluable correct responses have incomplete warrants, with each model's results averaged equally across supported task--probe combinations.
Among correct but warrant-incomplete responses, 35.9\% fail at least two process layers.
In distinguishing complete from incomplete warrants among correct answers, the best of seventeen existing evaluators achieves only 57.69\% task-averaged AUROC under the evaluation protocol.
Correct decisions thus frequently come with omissions or inconsistencies in their stated grounds, including failures spanning multiple process layers.
The tested evaluators also struggle to recognize these gaps.
We also train a simple reward model (\mbox{Warrant-RM}) on the benchmark labels, achieving 69.24\% warrant AUROC among correct answers versus 58.87\% for an outcome-supervised scorer with the same base model, inputs, and sample weights.
In downstream selection, Warrant-RM leads both outcome-supervised baselines in mean warrant completeness at every candidate count above one and in mean accuracy at most counts across both evaluation views.
\textbf{Together, these results show how RGDT-Bench connects reasoning diagnosis with warrant supervision to improve completeness assessment and response selection, supporting more reliable reasoning in RGDTs.}

\section{Related Work}
\label{sec:related_work}

\paragraph{Rule-based and evidence-grounded reasoning.}
Rule-governed decisions combine rule application with evidence support.
Benchmarks test natural-language rule inference \citep{clark2020ruletaker,tafjord2021proofwriter}, logically annotated entailment \citep{han2024folio}, and newly specified rules \citep{ma2025korbench}. Evidence verification tests passage support for claims \citep{thorne2018fever,wadden2020scifact}.
Policy and contract tasks connect these abilities through conditional answers and evidence-linked entailment \citep{sun2022conditionalqa,koreeda2021contractnli}.
RuleArena evaluates rule-use precision, recall, and application correctness \citep{zhou2025rulearena}, while SingGuard applies policy-conditioned rules to multimodal safety moderation \citep{singguard2026}.
These perspectives motivate \rgdt{}'s checks of required grounds and their combination into a decision.

\paragraph{Warrants and reasoning coverage.}
Assessing these explanations requires distinguishing stated support from factors that causally influenced the prediction \citep{jacovi2020faithfully,turpin2023unfaithful}.
SemEval-2018 Task~12 examines how implicit warrants connect premises to claims \citep{habernal2018semeval}.
For generated reasoning, \citet{emmons2025pragmatic} evaluate answer recovery without additional non-trivial reasoning and detection of synthetic omissions.
To check omissions in rule-governed decisions, \rgdt{} specifies reference requirements for rule use, conditions, evidence, and aggregation.

\paragraph{Process evaluation and supervision.}
Process supervision addresses reasoning beyond answer correctness.
Studies of correct mathematical solutions reveal errors in their reasoning \citep{uesato2022process}. Human and automated step labels provide more detailed supervision \citep{lightman2023lets,taur2024mathshepherd}.
ProcessBench and PRMBench assess error localization and fine-grained step judgments \citep{zheng2025processbench,song2025prmbench}, while multi-domain process reward models extend beyond mathematics \citep{zeng2025versaprm}.
Hard2Verify further examines whether mathematical steps have supported premises \citep{pandit2025hard2verify}, and the Thinking Reward Model (TRM) learns trace preferences from verified-correct solutions represented as graphs \citep{zhang2026trm}.
\rgdt{} complements these approaches with response-level completeness labels to train evaluators to recognize missing decision grounds.

\paragraph{Reward models and downstream selection.}
Reward supervision must also be evaluated through assessment and selection.
RewardBench, RM-Bench, and JudgeBench assess preferences, content/style sensitivity, and correctness \citep{lambert2025rewardbench,liu2025rmbench,tan2025judgebench}. VerifyBench extends this evaluation to reference-based rewards \citep{yan2026verifybench}.
Rubrics specify scoring criteria through prompts or reference-conditioned training \citep{liu2023geval,kim2023prometheus}, and can supervise policies through reasoning criteria or information coverage \citep{xie2026stepwise,wang2026refuse}.
However, persuasive instruction violations, length, and style can distort judgments \citep{zeng2023llmbar,dubois2024lengthcontrolled,feuer2025style,zheng2023llmjudge}.
To assess stated warrants, we separate model-assisted claim extraction from deterministic reference checks and evaluate scorers without exposing those references.
Beyond assessment, response selection can use verifier ranking or self-consistency \citep{cobbe2021verifiers,wang2023selfconsistency}, although gains depend on the problem and strategy \citep{snell2025testtime}.
\rgdt{} tests whether warrant supervision improves warrant assessment and selection over outcome supervision.

\section{Task and Warrant Evaluation}

\label{sec:task}

We define task inputs and generated responses, establish reference warrants, and derive completeness checks and the correct-answer warrant gap.

\subsection{Task Definition and Response Generation}

Rule-Governed Decision Tasks (RGDTs) require a justified decision given \(a=(q_a,F_a,R_a)\): a question \(q_a\), a fact set \(F_a\), and a rule set \(R_a\) expressed in natural language. Rules specify conditions \(c\), propositions about the case to assess, and how their judgments determine the decision. Facts used to assess whether \(c\) holds form its evidence set \(E_a(c)\subseteq F_a\). Our unified, extensible schema makes these rule--condition and condition--evidence links explicit.

\begin{figure*}[t]
  \centering
  \setlength{\abovecaptionskip}{4pt}
  \includegraphics[width=0.95\textwidth]{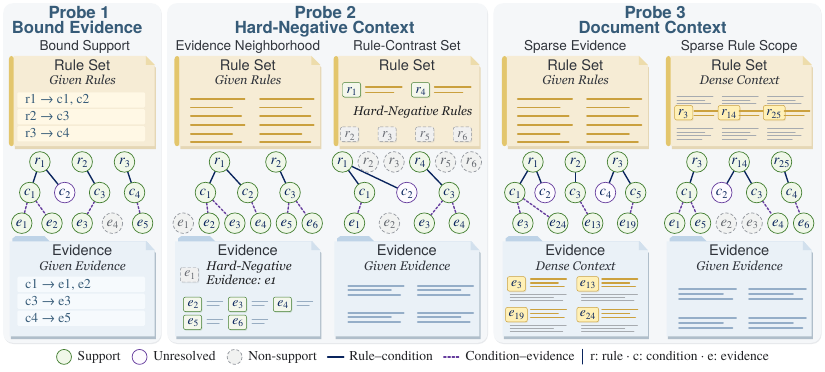}
  \caption{Warrant-access probes in RGDT-Bench. Circles denote rules ($r$), conditions ($c$), and evidence ($e$). Solid links connect rules to their conditions, and dashed links connect evidence to the conditions it supports. A rule may specify several conditions, and a condition may have multiple supporting passages or remain unresolved without support. P2/P3 illustrate evidence-dense and rule-dense contexts, with hard negatives in P2 and longer natural documents in P3. Support markings are illustrative and hidden from generators.}
  \label{fig:benchmark_design}
\end{figure*}

To solve instance \(a\), a \textbf{generator} produces a two-part response \(\tau=(h,\hat y)\): the \emph{native reasoning trace} \(h\) (a written chain-of-thought, CoT) and the separately reported \emph{terminal decision} \(\hat y\). The response's stated reasoning is its \emph{warrant}, whose completeness \rgdt{} assesses for diagnosis and supervision. The decision is correct when \(\hat y=y_a\), where \(y_a\) is the reference answer used for evaluation and withheld from the generator. Figure \ref{fig:benchmark_design} introduces three probes that vary support access to diagnose reasoning while preserving the decision problem (Section \ref{sec:probe_construction}).

\subsection{Reference Warrants}

To evaluate responses against source requirements, we use a \emph{reference warrant} \(w\), which records the required condition judgments, support, and decision logic. Structured records \((a,\tau,w)\) separate input, response, and reference requirements, keeping reference judgments and verification annotations hidden from the generator. This distinguishes what the response states from what it must justify.

These requirements depend on the task: question-level judgments for \emph{Policy}, inference conditions for \emph{Contract}, decision conditions for \emph{Regulation}, and identification of the proposed transaction plus its overall compliance for \emph{Transaction}. For each condition judgment, the reference specifies the applicable rules and required evidence. Evidence may comprise jointly required passages or accepted alternatives. Because the sources for \emph{Policy} and \emph{Transaction} do not list all evidence passages for each condition, we check whether the response's evidence supports its condition judgments, without requiring it to identify every supporting evidence passage (Table \ref{tab:source_reference_contracts}).

\subsection{Warrant Completeness and Correct-Answer Gaps}

\begin{figure*}[t]
  \centering
  \includegraphics[width=0.95\textwidth]{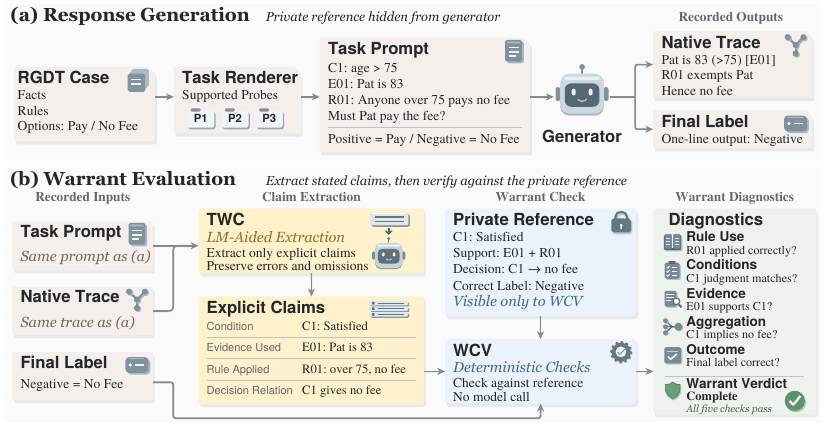}
  \caption{Response collection and diagnosis in RGDT-Bench. (a) A generator samples a native CoT and terminal decision from the RGDT input. (b) Using the outputs of (a), LLM-driven TWC extracts claims from the sampled CoT. WCV checks these claims and the terminal decision against the reference warrant, yielding condition-level diagnostics and a response-level completeness label.}
  \label{fig:diagnostic_pipeline}
\end{figure*}

Using these reference requirements, we check the response at four process layers and an outcome layer (Figure \ref{fig:diagnostic_pipeline}(b); Appendix \ref{app:task_details}). \textbf{(1) Rule use} checks whether each condition judgment uses the required governing rules: all jointly required passages, or one accepted alternative. \textbf{(2) Condition} checks whether every required judgment is stated and agrees with the reference, including unresolved conditions. \textbf{(3) Evidence} checks whether those judgments have the required support in the supplied materials, so merely citing a passage is insufficient. \textbf{(4) Aggregation} checks whether all reference-required decision conditions are combined according to the prescribed logic, such as requiring all conditions to hold or accepting alternatives. Thus, rule use concerns the grounds for individual judgments, whereas aggregation concerns their coverage and combination. The derived decision must agree with any decision stated in the trace and with the terminal decision. \textbf{(5) Outcome} checks the terminal decision against the reference answer. For \emph{Transaction} P2/P3, an additional \emph{rule-scope} diagnostic compares identified and reference rule sets without affecting completeness labels.

We define \emph{warrant completeness} as passing all five checks under at least one reference warrant. For response \(\tau\) to instance \(a\), the binary label \(W_{\mathrm{op}}(\tau,a)\) is 1 for completeness and 0 for an \emph{incomplete warrant}. A warrant can be incomplete even when every stated step is correct. Formally,

\begin{equation}
W_{\mathrm{op}}(\tau,a)=
\bigvee_{w\in\mathbb W(a)}
\bigwedge_{\ell\in\mathcal L_{\mathrm{op}}}
R_\ell(\tau,w),
\label{eq:operational_warrant}
\end{equation}
where \(\mathbb W(a)\) contains the reference warrants, \(\mathcal L_{\mathrm{op}}\) the five checks, and \(R_\ell(\tau,w)\) equals 1 if \(\tau\) passes check \(\ell\) under \(w\), and 0 otherwise.

Using these completeness labels, we measure how often correct answers lack required support. We define the \emph{correct-answer warrant gap} (CWG) as the proportion of evaluable correct responses with incomplete warrants. Missing required claims count as incompleteness, whereas extraction or parsing failures are not evaluable and are excluded. Writing \(C=1\) for evaluability, the gap is

\begin{equation}
\mathrm{CWG}=\Pr\!\left(W_{\mathrm{op}}(\tau,a)=0
\mid\hat y=y_a,\ C=1\right).
\label{eq:conditional_incomplete}
\end{equation}

Because it is computed within correct responses, this gap is not the difference between overall answer accuracy and warrant completeness.

\section{Benchmark Construction and Validation}

\label{sec:benchmark}

Following the RGDT formulation, RGDT-Bench covers policy applicability, contract inference, administrative-rule application, and basketball-transaction compliance. Three probes vary support presentation to diagnose reasoning with matched support, hard negatives, and natural documents. We then sample and annotate responses, validate the labels, and form balanced splits.

\subsection{Warrant-Access Probe Design}

\label{sec:probe_construction}

We organize case facts and rules into the three input settings shown in Figure \ref{fig:benchmark_design}: \textbf{P1: Bound evidence} supplies condition-matched facts as evidence and the governing rules, leaving the generator to judge conditions and derive the decision without reference judgments. \textbf{P2: Hard-negative context} retains this support and adds selected passages that resemble relevant evidence or rules but do not determine the conditions, testing whether the generator can distinguish useful support from misleadingly similar material. \textbf{P3: Document context} retains longer excerpts of natural contracts or rule documents, testing whether the generator can locate and apply support amid passages and document structure.

All four tasks support P1. We extend \emph{Contract} and \emph{Transaction} to P2/P3 because their contracts and rule manuals provide passages for hard negatives and document context. In contrast, \emph{Policy} and \emph{Regulation} organize support around question-specific evidence and rule snippets, so we retain P1 to preserve that context. This yields eight task--probe combinations and 904 prompts. Construction preserves complete source passages, their order, and the reference decision (Appendix \ref{app:data_construction}).

For comparison on the same case, P2 adds hard negatives to its P1 materials without changing its support or answer. P3 retains a longer support-containing contract excerpt for \emph{Contract} and expands P2 with surrounding rule-manual passages for \emph{Transaction} (Appendix \ref{app:input_conditions}). Because this changes information location, density, and context alongside length \citep{liu2024lost}, P3 tests document-context reasoning, where added context may clarify support or distract.

\subsection{Response Sampling and Automatic Annotation}

\label{sec:information_barriers}

To obtain responses for diagnosis and supervision, we use these probe inputs to prompt six generators from the Qwen, gpt-oss, and DeepSeek families for 16 responses per prompt, yielding 86,784 trace--decision pairs. Automatic annotation separates \emph{Claim Extraction} from \emph{Warrant Check} (Figure \ref{fig:diagnostic_pipeline}), first extracting stated reasoning and then assessing it against reference warrants.

\paragraph{Claim Extraction.}
An LLM-driven Trace Warrant Canonicalizer (TWC) structures the response's stated rule use, condition judgments, evidence, and aggregation without filling gaps or correcting errors. To prevent leakage, it receives task inputs, traces, and condition descriptions but not reference annotations or the separate terminal decision. Explicit paraphrases can identify passages without citation identifiers. The extracted claims then provide structured input for warrant checking.

\paragraph{Warrant Check.}
Using these extracted claims, a deterministic Warrant Consistency Verifier (WCV) checks their support and the terminal decision against the reference warrant. This extends evidence-support evaluation \citep{gao2023alce,li2024attributionbench} to Section \ref{sec:task}'s five checks: rule use, condition, evidence, aggregation, and outcome. Labels cover 86,435 responses and 202.1K condition judgments, providing fine-grained warrant supervision for evaluator assessment and reward-model training.

\subsection{Label Validation and Balanced Splits}

To validate these labels, we manually review 360 responses across predefined strata, obtaining \(84.2\%\) agreement (Krippendorff's \(\alpha=0.687\)), with \(92.7\%\) precision and \(78.5\%\) recall for complete warrants (Appendix \ref{app:validity_audits}). Despite residual errors, labels supply decision-ground supervision beyond answer correctness. Sections \ref{sec:rq4} and \ref{sec:rq5} test its value for assessment and response selection.

To use these labels without data leakage, we group related documents, operations, and similar cases, keeping all probe variants and responses within one split. To avoid overrepresenting one family's response style, we balance annotated responses (Section \ref{sec:information_barriers}) across six generators, with a smaller and a larger model from each of three families. We also balance tasks and warrant labels, with nearly equal probe coverage and attention to response length (Table \ref{tab:app_v16_balanced_collections}). Natural distributions support prevalence and attribution, while balanced subsets support evaluator comparison and training. Data organization precedes scoring, and reference annotations serve only as supervision and evaluation targets. These choices support accurate diagnosis and high-quality warrant supervision.

\raggedbottom
\section{Empirical Evaluation}
\label{sec:experiments}
\label{sec:evaluation_protocol}

\begin{wraptable}[18]{r}{0.46\textwidth}
\vspace{-6pt}
\setlength{\abovecaptionskip}{0pt}
\setlength{\belowcaptionskip}{4pt}
\caption{Correct-answer warrant gaps by generator (\%). Final/Warr.: answer accuracy/warrant completeness. \(\mathrm{CWG}'\): gap without the aggregation check. Equal-weight averages over supported task--probe combinations.}
\label{tab:rq1_prevalence}
\centering
\small
\renewcommand{\arraystretch}{1.05}
\setlength{\tabcolsep}{1.5pt}
\setlength{\rgdttableedgepad}{3pt}
\newlength{\rqonecolwidth}
\setlength{\rqonecolwidth}{\dimexpr(\linewidth-0.20\linewidth-8\tabcolsep-2\rgdttableedgepad-\arrayrulewidth)/4\relax}
\begin{tabular}{@{\hspace{\rgdttableedgepad}}>{\raggedright\arraybackslash}m{0.20\linewidth}|*{4}{>{\centering\arraybackslash}m{\rqonecolwidth}}@{\hspace{\rgdttableedgepad}}}
\toprule
\tablehead{Size} & \tablehead{Final $\uparrow$} &
\tablehead{Warr. $\uparrow$} & \tablehead{CWG $\downarrow$} &
\tablehead{\(\mathrm{CWG}'\) $\downarrow$} \\
\midrule
\tablegroup{5}{DeepSeek-R1-Distill-Llama}
\rgdtlogodeepseek{}8B  & 43.9 & 10.7 & 76.7 & 73.6 \\
\rgdtlogodeepseek{}70B & 49.1 & 26.1 & 48.8 & 46.6 \\
\specialrule{\lightrulewidth}{0pt}{0pt}
\tablegroup{5}{gpt-oss}
\rgdtlogoopenai{}20B  & 74.4 & 47.9 & 36.7 & 32.2 \\
\rgdtlogoopenai{}120B & 77.7 & 56.3 & 28.9 & 22.1 \\
\specialrule{\lightrulewidth}{0pt}{0pt}
\tablegroup{5}{Qwen3.5}
\rgdtlogoqwen{}9B  & 83.8 & 63.5 & 25.4 & 22.2 \\
\rgdtlogoqwen{}27B & 87.7 & 66.6 & 24.8 & 17.5 \\
\midrule
\tabletotalrow
\textbf{Mean} & \textbf{69.4} & \textbf{45.2} & \textbf{40.2} & \textbf{35.7} \\
\bottomrule
\end{tabular}
\end{wraptable}

Using RGDT-Bench, we first establish how often correct answers have incomplete warrants and attribute these failures to reasoning layers. We then use existing \textbf{evaluators}, comprising LLM judges and reward models (RMs), to score responses without reference annotations and test whether they distinguish complete from incomplete warrants. Their limitations motivate training RMs with warrant supervision, whose value we assess first in completeness assessment and then in downstream response selection. Together, these results show how RGDT-Bench connects the diagnosis of reasoning failures to supervision that improves both warrant assessment and response selection.

\setlength{\columnsep}{10pt}
\setlength{\intextsep}{2pt}
\Needspace*{40pt}
\subsection{RQ1: How Often Do Correct Answers Have Incomplete Warrants?}
\label{sec:rq1}

We first examine whether correct decisions are accompanied by complete warrants. CWG quantifies the proportion of evaluable correct responses with incomplete warrants (Equation \ref{eq:conditional_incomplete}). Table \ref{tab:rq1_prevalence} shows that CWG ranges from 24.8\% to 76.7\% across six generators and that the larger model in each family has higher warrant completeness. Yet family-level gaps remain substantial (Figure \ref{fig:rq12_compact_summary}a), showing that stronger overall performance does not ensure that correct decisions are justified by complete warrants.

To test whether explicit aggregation requirements explain these gaps, we ignore aggregation failures while retaining all other checks. Table \ref{tab:rq1_prevalence} reports the relaxed gap as \textbf{\(\mathrm{CWG}'\)}, retaining rule-use, condition, and evidence failures (Appendix \ref{app:aggregation_waiver}). Even under this relaxed check, mean CWG falls only from 40.2\% to 35.7\%: \textbf{incomplete warrants remain widespread among correct answers, posing potential safety risks in RGDTs.}
\par
\RGDTfinishwrap
\Needspace*{242pt}
\subsection{RQ2: How Are Warrant Gaps Distributed Across Reasoning Layers?}
\label{sec:rq2}

\setlength{\columnsep}{9pt}
\setlength{\intextsep}{3pt}
\begin{wrapfigure}[21]{r}{0.50\textwidth}
  \centering
  \includegraphics[width=\linewidth]{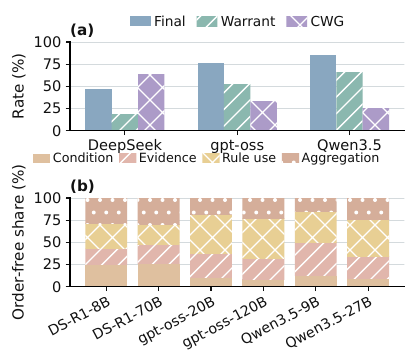}
  \caption{(a) Family-equal averages of Final (answer accuracy), Warrant (warrant completeness), and CWG (correct-answer warrant gap). (b) Each correct but incomplete response contributes equally to its failed layers. DS-R1 denotes DeepSeek-R1-Distill-Llama.}
  \label{fig:rq12_compact_summary}
  \label{fig:aggregate_paired_summary}
  \label{fig:rq2_failure_attribution}
\end{wrapfigure}

To understand the incomplete warrants observed in Section \ref{sec:rq1}, we examine four process layers: rule use, condition, evidence, and aggregation. Failures overlap: 35.9\% of incomplete warrants fail at least two requirements (Appendix Table \ref{tab:rq2_failure_anatomy}). To give each response equal weight, we divide its unit contribution among its failed layers. For layers \(d\in\mathcal D\), let \(F_{id}=1\) if response \(i\) fails at \(d\), and 0 otherwise. Each failed layer receives a share \(\phi_{id}\) equal to one divided by the number of failed layers, while other layers receive zero:
\begin{equation}
 \phi_{id}=\frac{F_{id}}{\sum_{d'\in\mathcal D}F_{id'}},\qquad
 \sum_{d\in\mathcal D}\phi_{id}=1,
 \label{eq:order_free_failure_share}
\end{equation}

Averaging these shares yields the failure profiles in Figure \ref{fig:rq12_compact_summary}b. Rule use remains the largest contributor under alternative task weights and P1-only evaluation, with aggregation contributing more than evidence (Appendix Table \ref{tab:rq2_weight_sensitivity}). This variation motivates matched-case probe comparisons to examine how support presentation affects failures (Appendix \ref{app:probe_effect_readout}). Since failures overlap, these profiles locate weaknesses rather than isolate causes. \textbf{Together, these analyses show how RGDT-Bench supports layered diagnosis of models' reasoning weaknesses in RGDTs across input conditions, beyond answer accuracy.}

\subsection{RQ3: How Well Do Existing Evaluators Assess Warrant Completeness?}
\label{sec:rq3}

Since correct answers often have incomplete warrants, we compare fourteen LLM judges and three pretrained scoring models on distinguishing complete from incomplete warrants: Skywork-Reward-V2-Llama-3.1-8B \citep{liu2026skyworkrewardv2}, dORM \citep{lee2025rethinkingrewardmodels}, and TRM \citep{zhang2026trm}. All receive prompts, traces, and decisions without reference warrants.

\begin{table}[!t]
\caption{Existing evaluators' warrant assessment (\%). Positive class: complete warrants. Random baselines: AUROC/BalAcc, 50\%; AUPRC, 50.00\%/63.76\% (full/correct-only). Bold/underline: best/second.}
\label{tab:main_evaluator_utility}
\centering
\small
\renewcommand{\arraystretch}{1.20}
\setlength{\heavyrulewidth}{0.7pt}
\setlength{\lightrulewidth}{0.4pt}
\setlength{\cmidrulewidth}{0.35pt}
\setlength{\tabcolsep}{1pt}
\settowidth{\dimen0}{\rgdtlogoskywork{}Skywork-Reward-V2-Llama-3.1-8B}
\edef\rgdtmetricwidth{\the\dimexpr(\textwidth-\dimen0-6pt-11\tabcolsep-2\rgdttableedgepad-2\arrayrulewidth)/6\relax}
\begin{tabular}{@{\hspace{\rgdttableedgepad}}l@{\hspace{6pt}}|*{3}{>{\centering\arraybackslash}m{\rgdtmetricwidth}}|*{3}{>{\centering\arraybackslash}m{\rgdtmetricwidth}}@{\hspace{\rgdttableedgepad}}}
\toprule
\tableheaderrow
\multirow{2}{*}{\tablehead{Evaluator}} &
\multicolumn{3}{c|}{\tablehead{Full evaluation view}} &
\multicolumn{3}{c}{\tablehead{Correct-only view}} \\
\cmidrule(lr){2-4}\cmidrule(lr){5-7}
\tableheaderrow
& \makebox[\linewidth][c]{\footnotesize\tablehead{AUROC $\uparrow$}} & \makebox[\linewidth][c]{\footnotesize\tablehead{AUPRC $\uparrow$}}
& \makebox[\linewidth][c]{\footnotesize\tablehead{BalAcc $\uparrow$}} & \makebox[\linewidth][c]{\footnotesize\tablehead{AUROC $\uparrow$}}
& \makebox[\linewidth][c]{\footnotesize\tablehead{AUPRC $\uparrow$}} & \makebox[\linewidth][c]{\footnotesize\tablehead{BalAcc $\uparrow$}} \\
\midrule
\tablegroup{7}{Other LLM Judges}
\rgdtlogoopenai{}GPT-5.6 Sol & \tablebest{70.19} & \tablesecond{63.84} & \tablebest{67.09} & 53.44 & 66.14 & 53.51 \\
\rgdtlogogoogle{}Gemini-2.5-Pro & 65.37 & 59.10 & 65.62 & 53.83 & 65.61 & 53.82 \\
\rgdtlogozhipu{}GLM-5.2 & 64.07 & 58.20 & \tablesecond{66.05} & 49.00 & 63.02 & \tablesecond{54.43} \\
\specialrule{\lightrulewidth}{1pt}{1pt}
\tablegroup{7}{Qwen3.5 Series}
\rgdtlogoqwen{}Qwen3.5-0.8B & 59.53 & 56.14 & 58.29 & 53.40 & 65.95 & 52.43 \\
\rgdtlogoqwen{}Qwen3.5-4B & 68.49 & 62.13 & 63.92 & \tablebest{57.69} & 67.92 & 53.10 \\
\rgdtlogoqwen{}Qwen3.5-9B & 68.18 & 61.57 & 64.97 & 56.17 & 66.74 & 54.36 \\
\rgdtlogoqwen{}Qwen3.5-27B & 66.86 & 60.73 & 64.97 & 52.64 & 65.08 & 53.40 \\
\rgdtlogoqwen{}Qwen3.5-35B-A3B & 66.92 & 60.24 & 65.31 & 54.80 & 66.04 & 54.01 \\
\specialrule{\lightrulewidth}{1pt}{1pt}
\tablegroup{7}{gpt-oss Series}
\rgdtlogoopenai{}gpt-oss-20B & 64.82 & 59.25 & 62.40 & 53.84 & 65.55 & 53.05 \\
\rgdtlogoopenai{}gpt-oss-120B & \tablesecond{68.83} & 63.06 & 64.04 & \tablesecond{57.47} & \tablesecond{68.01} & 53.70 \\
\specialrule{\lightrulewidth}{1pt}{1pt}
\tablegroup{7}{DeepSeek-R1 Distillation Family}
\rgdtlogodeepseek{}DeepSeek-R1-Distill-Llama-8B & 57.83 & 55.17 & 58.03 & 53.42 & 65.95 & 53.92 \\
\rgdtlogodeepseek{}DeepSeek-R1-Distill-Llama-70B & 66.40 & 60.23 & 63.91 & 55.48 & 66.55 & 54.14 \\
\specialrule{\lightrulewidth}{1pt}{1pt}
\tablegroup{7}{Llama Series}
\rgdtlogometa{}Llama-3.1-8B-Instruct & 62.13 & 57.33 & 61.60 & 49.94 & 63.94 & 53.11 \\
\rgdtlogometa{}Llama-3.1-70B-Instruct & 67.54 & 61.45 & 63.89 & 55.77 & 67.03 & 54.39 \\
\specialrule{\lightrulewidth}{1pt}{1pt}
\tablegroup{7}{Pretrained Scoring Models}
\rgdtlogoskywork{}Skywork-Reward-V2-Llama-3.1-8B & 63.31 & 58.09 & 63.24 & 51.89 & 64.77 & \tablebest{55.84} \\
\rgdtlogodormscore{}dORM-8B & 67.48 & \tablebest{65.56} & 60.58 & 54.37 & \tablebest{69.36} & 50.66 \\
\rgdtlogotrmtrace{}TRM-8B & 61.45 & 59.42 & 56.25 & 54.05 & 66.80 & 51.99 \\
\bottomrule
\end{tabular}
\end{table}

To separate this ability from answer correctness, we use the \textbf{full evaluation view} (label-balanced data) and the \textbf{correct-answer-only view} (its correct responses). AUROC and average precision (AUPRC) measure discrimination, while balanced accuracy (BalAcc) measures classification. AUPRC comparisons stay within each view because prevalence differs. Table \ref{tab:main_evaluator_utility} averages each metric \(m_{tp}\) for task \(t\) and probe \(p\) first over supported probes \(\mathcal P_t\), then equally across the four tasks \(\mathcal T\):

\vspace{-5pt}
\begingroup
\setlength{\abovedisplayskip}{5pt}
\setlength{\belowdisplayskip}{5pt}
\setlength{\abovedisplayshortskip}{3pt}
\setlength{\belowdisplayshortskip}{5pt}
\begin{equation}
 \overline m_{\mathrm{task}}=
 \frac{1}{|\mathcal T|}\sum_{t\in\mathcal T}
 \frac{1}{|\mathcal P_t|}\sum_{p\in\mathcal P_t}m_{tp}.
 \label{eq:task_equal_estimand}
\end{equation}
\par\endgroup
\vspace{-5pt}

Table \ref{tab:main_evaluator_utility} shows that although GPT-5.6 Sol leads the full view with 70.19\% AUROC, it reaches only 53.44\% (random: 50\%) among correct answers, where the best existing evaluator reaches 57.69\%. Llama-3.1-70B detects just 10.77\% of incomplete warrants despite 98.01\% complete-warrant recall (Appendix Table \ref{tab:rq3_class_recall}). \textbf{These results show that existing evaluators struggle to assess warrant completeness in correct RGDT answers, motivating explicit warrant supervision.}

\Needspace*{75pt}
\subsection{RQ4: Can Warrant Supervision Improve Completeness Assessment?}
\label{sec:rq4}

To address these assessment difficulties, we train three Qwen3.5-9B reward models as evaluators under matched settings (Appendix \ref{app:training_evaluation_protocol}). Final-only ORM receives the prompt and decision, while Full-response ORM also receives the trace. Both learn correctness scores from outcome labels. Warrant-RM uses the same full-response inputs but learns completeness scores from warrant labels.

\begin{table}[!t]
\caption{Effects of trace input and supervision on RM assessment (\%). Scores report the mean \(\pm\) standard deviation across three training runs. Views include all responses or only correct answers. Outcome metrics are N/A in the correct-only view because it contains one outcome class. Bold/underline: best/second.}
\label{tab:main_downstream_utility_full}
\centering
\small
\renewcommand{\arraystretch}{1.20}
\setlength{\heavyrulewidth}{0.7pt}
\setlength{\lightrulewidth}{0.4pt}
\setlength{\cmidrulewidth}{0.35pt}
\setlength{\tabcolsep}{2.5pt}
\newlength{\rgdtdstablewidth}
\setlength{\rgdtdstablewidth}{0.88\textwidth}
\newlength{\rgdtdsmodelwidth}
\setlength{\rgdtdsmodelwidth}{72pt}
\newlength{\rgdtdsmetricwidth}
\setlength{\rgdtdsmetricwidth}{\dimexpr(\rgdtdstablewidth-\rgdtdsmodelwidth-12\tabcolsep-2\rgdttableedgepad-2\arrayrulewidth)/6\relax}
\begin{tabular}{@{\hspace{\rgdttableedgepad}}>{\raggedright\arraybackslash}m{\rgdtdsmodelwidth}|*{3}{>{\centering\arraybackslash}m{\rgdtdsmetricwidth}}|*{3}{>{\centering\arraybackslash}m{\rgdtdsmetricwidth}}@{\hspace{\rgdttableedgepad}}}
\toprule
\tableheaderrow
\multirow{2}{*}{\tablehead{Model}} &
\multicolumn{3}{c|}{\tablehead{Full evaluation view}} &
\multicolumn{3}{c}{\tablehead{Correct-only view}} \\
\cmidrule(lr){2-4}\cmidrule(lr){5-7}
\tableheaderrow
& \mbox{\tablehead{AUROC $\uparrow$}} & \mbox{\tablehead{AUPRC $\uparrow$}} & \mbox{\tablehead{BalAcc $\uparrow$}}
& \mbox{\tablehead{AUROC $\uparrow$}} & \mbox{\tablehead{AUPRC $\uparrow$}} & \mbox{\tablehead{BalAcc $\uparrow$}} \\
\midrule
\tablegroup{7}{Warrant Completeness}
Final-only ORM & $67.31\sd{2.06}$ & $64.38\sd{2.59}$ & $64.50\sd{2.23}$ & $51.86\sd{3.40}$ & $66.91\sd{2.65}$ & $51.84\sd{1.39}$ \\
Full-response ORM & $\underline{72.99}\sd{2.56}$ & $\underline{68.96}\sd{4.91}$ & $\underline{67.46}\sd{0.43}$ & $\underline{58.87}\sd{4.61}$ & $\underline{71.20}\sd{5.03}$ & $\underline{53.86}\sd{1.21}$ \\
Warrant-RM & $\mathbf{79.93}\sd{1.85}$ & $\mathbf{79.12}\sd{1.52}$ & $\mathbf{70.59}\sd{1.88}$ & $\mathbf{69.24}\sd{2.15}$ & $\mathbf{80.79}\sd{1.11}$ & $\mathbf{58.86}\sd{1.41}$ \\
\specialrule{\lightrulewidth}{1pt}{1pt}
\tablegroup{7}{Outcome Correctness}
Final-only ORM & $89.55\sd{2.31}$ & $96.24\sd{0.85}$ & $\underline{83.87}\sd{2.61}$ & N/A & N/A & N/A \\
Full-response ORM & $\underline{91.48}\sd{0.92}$ & $\underline{96.54}\sd{0.23}$ & $\mathbf{84.75}\sd{1.46}$ & N/A & N/A & N/A \\
Warrant-RM & $\mathbf{91.91}\sd{0.91}$ & $\mathbf{97.06}\sd{0.51}$ & $81.48\sd{1.32}$ & N/A & N/A & N/A \\
\bottomrule
\end{tabular}
\end{table}

\par

\Needspace*{180pt}
\setlength{\columnsep}{10pt}
\setlength{\intextsep}{4pt}
\begin{wrapfigure}[15]{r}{0.50\textwidth}
  \vspace{-7pt}
  \centering
  \includegraphics[width=\linewidth]{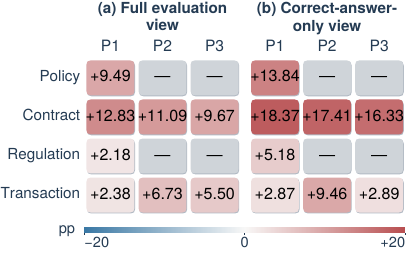}
  \caption{Warrant-supervision AUROC gains (pp). Cells show mean Warrant-RM minus Full-response ORM over three fits in both views. Gray: unsupported.}
  \label{fig:rq4_task_probe_gains}
\end{wrapfigure}
To compare these models consistently, we use Section \ref{sec:rq3}'s views and metrics (Table \ref{tab:main_downstream_utility_full}). \textbf{(1) Trace access.} Full-response ORM improves warrant AUROC over Final-only ORM by 5.68 pp in the full view and 7.01 pp among correct answers. \textbf{(2) Warrant supervision.} With identical inputs, Warrant-RM improves AUROC over Full-response ORM by 6.94 pp in the full view and 10.37 pp among correct answers. Correct-only gains are positive across tasks and supported task--probe combinations (Figure \ref{fig:rq4_task_probe_gains}). \textbf{(3) Comparison with existing evaluators.} Warrant-RM gains 11.55 pp in correct-only AUROC over Section \ref{sec:rq3}'s best evaluator.

\par

To test whether these AUROC gains also help identify incomplete warrants, we apply score thresholds. Among correct answers, Warrant-RM increases incomplete-warrant detection from \(20.69\%\) to \(33.56\%\) relative to Full-response ORM, while complete-warrant recall changes from \(87.02\%\) to \(84.16\%\) (Appendix Table \ref{tab:rq3_class_recall}). This improves gap detection at the cost of some complete-warrant recall. \textbf{Together, these results show that warrant supervision improves completeness assessment beyond trace access and outcome supervision, motivating its use for downstream response selection.}

\par\RGDTfinishwrap

\raggedbottom
\Needspace*{40pt}
\subsection{RQ5: Do Learned Warrant Scores Improve Response Selection?}
\label{sec:rq5}

\setlength{\columnsep}{10pt}
\setlength{\intextsep}{5pt}
\begin{wraptable}[20]{r}{0.50\textwidth}
\centering
\caption{Warrant-supervision effects on response selection (\%). Means over three fits. Final/Full: Final-only/Full-response ORM; WRM: Warrant-RM. Bold/underline: best/second.}
\label{tab:bon_utility}
\footnotesize
\setlength{\tabcolsep}{1.5pt}
\renewcommand{\arraystretch}{1.08}
\begin{tabular}{@{\hspace{2pt}}r|*{3}{W{c}{26pt}}|*{3}{W{c}{26pt}}@{\hspace{2pt}}}
\toprule
 & \multicolumn{3}{c|}{\tablehead{Warrant $\uparrow$}} & \multicolumn{3}{c}{\tablehead{Accuracy $\uparrow$}} \\
\cmidrule(lr){2-4}\cmidrule(l){5-7}
\tablehead{$N$} & \tablehead{Final} & \tablehead{Full} & \tablehead{WRM} & \tablehead{Final} & \tablehead{Full} & \tablehead{WRM} \\
\midrule
\tablegroup{7}{All response groups}
1 & \textbf{51.01} & \textbf{51.01} & \textbf{51.01} & \textbf{72.58} & \textbf{72.58} & \textbf{72.58} \\
4 & 54.42 & \underline{55.90} & \textbf{56.92} & 78.21 & \textbf{78.39} & \underline{78.30} \\
8 & 55.34 & \underline{57.77} & \textbf{58.94} & 79.82 & \underline{80.41} & \textbf{80.55} \\
12 & 55.85 & \underline{58.79} & \textbf{59.94} & 80.49 & \underline{81.43} & \textbf{81.71} \\
16 & 56.30 & \underline{59.47} & \textbf{60.70} & 80.81 & \underline{82.04} & \textbf{82.50} \\
\midrule
\tablegroup{7}{Target-specific opportunity groups}
1 & \textbf{53.50} & \textbf{53.50} & \textbf{53.50} & \textbf{58.66} & \textbf{58.66} & \textbf{58.66} \\
4 & 58.33 & \underline{61.81} & \textbf{63.73} & 68.55 & \underline{70.50} & \textbf{70.79} \\
8 & 59.46 & \underline{65.42} & \textbf{67.57} & 70.81 & \underline{74.36} & \textbf{75.31} \\
12 & 60.04 & \underline{67.66} & \textbf{69.70} & 71.27 & \underline{76.33} & \textbf{77.58} \\
16 & 60.50 & \underline{69.34} & \textbf{71.52} & 71.12 & \underline{77.57} & \textbf{79.23} \\
\bottomrule
\end{tabular}
\end{wraptable}

To test whether better assessment improves response selection, we compare two views: \textbf{(1) All response groups} with all sixteen candidates and resolved labels; \textbf{(2) Target-specific opportunity groups} containing both positive and negative labels for the evaluated target. Selectors share groups within each view and choose the highest-scoring response without reference labels (Appendix \ref{app:score_independent_selection}). Full results for \(N=1\)--\(16\) appear in Appendix Table \ref{tab:bon_all_groups}.

Across both views, Warrant-RM achieves the highest mean completeness at every \(N>1\) and the highest mean accuracy at most candidate counts (Table \ref{tab:bon_all_groups}). On all response groups, completeness and accuracy improve by 4.40 and 1.69 pp over Final-only ORM, and by 1.23 and 0.46 pp over Full-response ORM (\(N=16\); rounded means in Table \ref{tab:bon_utility}). Within target-specific opportunity groups, Warrant-RM exceeds Full-response ORM by 2.18 pp in completeness and 1.66 pp in accuracy at \(N=16\), with a 1.99 pp mean completeness gain across \(N=2\)--\(16\). Mean gains persist with random tie-breaking (Appendix \ref{app:score_independent_selection}). \textbf{Together, these results show that even a simple warrant-supervised reward model improves completeness assessment and response selection.}

\RGDTfinishwrap

\setlength{\intextsep}{8pt plus 1pt minus 1pt}
\setcounter{topnumber}{3}

\section{Conclusion}
\label{sec:conclusion}

We formalize \textbf{Rule-Governed Decision Tasks (RGDTs)}, which require models to apply external rules to case facts and justify their decisions. Correct answers can conceal incomplete warrants, leaving potential safety risks unchecked. \rgdt{} combines a unified, extensible schema with four task families, controlled probes, and balanced evaluation. Its layered checks support joint diagnosis of reasoning failures, which existing evaluators struggle to recognize among correct answers. The resulting warrant supervision improves reward-model assessment and downstream response selection over outcome-supervised baselines. \rgdt{} thus connects failure diagnosis with learning signals for more reliable RGDT reasoning, enabling learning from such reasoning failures.

\section{Limitations and Future Work}
\label{sec:discussion}

Our evaluation has three limitations. First, decisions use supplied inputs, leaving multi-turn interaction and tool use untested. Future work can track how warrants evolve as evidence accumulates. Second, text-only inputs exclude visual evidence and document layout, motivating multimodal extensions that link these sources to condition judgments. Third, we study sequence-level reward modeling rather than step-level process reward models. Evaluating the latter requires consistent reasoning-step segmentation and alignment with condition-level supervision to compare process rewards in RGDTs.

\par\setlength{\parskip}{6pt}
\section*{AI Use Statement}

We used generative AI tools to assist with data synthesis, including candidate reasoning-trace generation, and with extracting stated claims during automatic annotation. Annotation quality was assessed through human review and agreement analysis (Appendix \ref{app:validity_audits}). We also used generative AI for language polishing and typesetting assistance. The authors take responsibility for the final content of this work, including text, claims, and artifacts produced with AI assistance.

\setlength{\parskip}{0pt}
\setlength{\bibsep}{0pt}
\renewcommand{\bibfont}{\fontsize{10}{10.5}\selectfont}
\bibliography{iclr2027_conference,references_trm}
\bibliographystyle{iclr2027_conference}

\clearpage
\setlength{\parskip}{6pt}
\appendix
\raggedbottom
\setlength{\textfloatsep}{10pt plus 5pt minus 2pt}
\setlength{\intextsep}{8pt plus 5pt minus 2pt}
\section*{Appendix Guide}
\label{app:guide}

The appendix follows the paper's logic from reference requirements and probe construction to annotation, evaluation, and supplementary evidence for warrant supervision. Prompt examples connect these stages through one case.

\begingroup
\normalsize
\setlength{\parindent}{0pt}
\newcommand{\appchapter}[3]{%
  \par\addvspace{9pt}%
  \noindent\makebox[\linewidth][l]{%
    \makebox[0.88\linewidth][l]{\hyperref[#2]{\textcolor{rgdtstructure}{\textbf{#1\quad #3}}}}%
    \makebox[0.12\linewidth][r]{\hyperref[#2]{\textcolor{rgdtmuted}{p.~\pageref*{#2}}}}}%
  \par}
\newcommand{\appitem}[3]{%
  \par\addvspace{7pt}%
  \noindent\makebox[\linewidth][l]{%
    \makebox[0.88\linewidth][l]{\hspace*{1.5em}\hyperref[#2]{\textcolor{rgdtstructure}{\textbf{#1}}\enspace #3}}%
    \makebox[0.12\linewidth][r]{\hyperref[#2]{\textcolor{rgdtmuted}{\pageref*{#2}}}}}%
  \par}

\newcommand{\appsubitem}[3]{%
  \par\addvspace{4pt}%
  \noindent\makebox[\linewidth][l]{%
    \makebox[0.88\linewidth][l]{\hspace*{3em}\hyperref[#2]{\textcolor{rgdtstructure}{#1}\enspace #3}}%
    \makebox[0.12\linewidth][r]{\hyperref[#2]{\textcolor{rgdtmuted}{\pageref*{#2}}}}}%
  \par}

\appchapter{A}{app:task_details}{Reference Warrants and Diagnostic Checks}

\appchapter{B}{app:data_construction}{Task Coverage, Data Balance, and Warrant-Access Probes}
\appitem{B.1}{app:source_provenance}{Task Coverage and Leakage Prevention}
\appitem{B.2}{app:balanced_collections}{Natural and Balanced Response Collections}
\appitem{B.3}{app:input_conditions}{Construction of the Three Warrant-Access Probes}

\appchapter{C}{app:barriers}{Automatic Annotation and Human Validation}
\appitem{C.1}{app:access_contract}{Information Boundaries}
\appitem{C.2}{app:twc_wcv_modules}{Claim Extraction and Warrant Verification}
\appitem{C.3}{app:validity_audits}{Human Review of Annotation Quality}

\appchapter{D}{app:experiments}{Evaluation Measures and Matched Comparisons}
\appitem{D.1}{app:metrics}{Metrics, Weighting, and Uncertainty}
\appitem{D.2}{app:study_design}{Existing Evaluators}
\appitem{D.3}{app:training_evaluation_protocol}{Matched Reward-Model Training}

\appchapter{E}{app:supplementary_results}{Supplementary Evidence for Diagnosis and Supervision}
\appitem{E.1}{app:population_sensitivity}{Robustness of Correct-Answer Warrant Gaps}
\appitem{E.2}{app:rq2_supplement}{Layered Attribution and Probe Comparisons}
\appsubitem{E.2.1}{app:layered_failure_attribution}{Layered Failure Attribution}
\appsubitem{E.2.2}{app:probe_effect_readout}{Matched Probe Comparisons}
\appitem{E.3}{app:rq3_supplement}{Probability Quality and Incomplete-Warrant Detection}
\appitem{E.4}{app:rq4_supplement}{Effects of Trace Access and Warrant Supervision}
\appsubitem{E.4.1}{app:final_correct_assessment}{Warrant Assessment Among Correct Answers}
\appsubitem{E.4.2}{app:shared_weight_ablation}{Effect of Training Weights}
\appitem{E.5}{app:score_independent_selection}{Response Selection and Robustness Checks}

\appchapter{F}{app:prompts}{Prompt Examples for Generation, Annotation, and Evaluation}
\endgroup
\raggedbottom

\clearpage
\FloatBarrier
\section{Reference Warrants and Diagnostic Checks}
\label{app:task_details}
\label{app:task_formalization}
\label{app:diagnostic_taxonomy}

To assess a response's warrant, the reference warrant specifies the required judgments, their support, and their roles in the decision. This appendix expands Section \ref{sec:task} by specifying support requirements, distinguishing rule use from aggregation, and explaining how unresolved checks are handled.

\paragraph{Task inputs and hidden references.}
The input schema separates the natural-language question, case facts, and rules. Evidence records link facts to the conditions they help assess, while reference judgments remain hidden from the generator. Reference warrants specify the required judgments, support, and decision logic. Completeness requires covering these grounds consistently, even when fewer stated grounds suffice for the decision.

\paragraph{Checking the support for each judgment.}
Rule-use and evidence checks ask whether each condition judgment has the required support. Rule use checks the governing rules, while evidence checks the supporting case information. For rules and evidence separately, the reference requires all specified passages or an accepted alternative. Evidence cannot replace required rule use, or vice versa. For example, if two evidence passages are jointly required, using only one is insufficient. If either passage is accepted as an alternative, using one meets this coverage requirement.

Some sources do not specify supporting passages for each judgment. In these cases, we do not require a particular preassigned passage, but still assess the support stated in the response against the task-specific requirements in Table \ref{tab:source_reference_contracts}. These checks concern the grounds for individual judgments. How those judgments combine into a decision is checked separately under aggregation.

\paragraph{From condition judgments to warrant completeness.}
Table \ref{tab:warrant_layers} groups the checks used by WCV. Rule use concerns the grounds for each judgment, whereas aggregation concerns whether all reference-required decision conditions are covered and combined correctly. The decision obtained by combining the stated condition judgments must agree with any decision stated in the trace and with the separately reported terminal decision. All five checks must pass against the same reference warrant. For \emph{Transaction}, we separately check that the response identifies the transaction action posed in the question and that its compliance judgment has the required grounds (Table \ref{tab:source_reference_contracts}). Compliance requires an accepted governing rule and meaningful use of the supplied evidence or rules.

\begin{table}[!htbp]
\caption{Warrant checks and separate diagnostics. The five completeness checks must pass. rule scope is a separate diagnostic. Failures shown are illustrative and may co-occur. $C=1$ denotes evaluability in Equation \ref{eq:conditional_incomplete}.}
\label{tab:warrant_layers}
\centering
\small
\renewcommand{\arraystretch}{1.14}
\setlength{\tabcolsep}{4pt}
\begin{tabular}{@{\hspace{\rgdttableedgepad}}
  >{\raggedright\arraybackslash}p{0.17\textwidth}
  >{\raggedright\arraybackslash}p{0.51\textwidth}
  >{\raggedright\arraybackslash}p{0.25\textwidth}@{\hspace{\rgdttableedgepad}}}
\toprule
\tableheaderrow
\tablehead{Layer} & \tablehead{Required property} & \tablehead{Representative failure} \\
\midrule
Rule scope & applied rules match a reference set where the comparison is determined, with unidentified applied rules marked NA & rule outside the reference set \\
\midrule
Rule use & each condition judgment has its required rule support, using all required rule passages or an accepted alternative & missing required rule use \\
Condition & every required condition receives the reference judgment & omitted exception condition \\
Evidence & semantic support satisfies the source-specific requirement independently of literal citation format & evidence does not support the judgment \\
Aggregation & all required decision conditions enter aggregation, with their judgments combined according to the decision logic and consistent with the terminal decision & omitted aggregation input or inconsistent decision \\
Outcome & predicted label equals the final answer implied by the same reference warrant & final answer wrong \\
\midrule
Evaluability \(C\) & stated claims can be extracted into the required structure, although missing claims still fail & failed extraction or non-parseable trace \\
\bottomrule
\end{tabular}
\end{table}

\paragraph{Separate diagnostics and evaluability.}
Rule scope separately compares identified and reference rule sets for \emph{Transaction} P2/P3 without affecting warrant completeness. When the applied rules cannot be identified, rule scope is marked NA (not available). Missing or incorrect claims fail the corresponding checks, whereas only extraction or parsing failures yield \textsc{not-evaluable}. Several checks may fail on the same response. The results identify missing or inconsistent grounds without claiming that one failure caused another.

\section{Task Coverage, Data Balance, and Warrant-Access Probes}
\label{app:data_construction}

RGDT-Bench covers four task tracks through a shared representation of decisions and their required grounds. To support both diagnosis and learning, its construction preserves task facts and rules, prevents related cases from crossing data splits, and varies access to support through three probes.

\subsection{Task Coverage and Leakage Prevention}
\label{app:source_provenance}
\label{app:split_lineage}

Cases sharing source documents, involving related operations, or having similar content are placed in the same source group. To prevent data leakage, each source group stays within one data split before probe construction or response generation. Table \ref{tab:source_provenance} reports source-case counts by split, with 90 test cases forming 76 such source groups.

\begin{table}[H]
\caption{Source-case counts by task and split. Each case yields one prompt per supported probe (checkmarks), with all probes sharing the case split. Native labels are balanced within each supported task--split--probe combination.}
\label{tab:source_provenance}
\label{tab:app_sample_balance}
\centering
\small
\renewcommand{\arraystretch}{1.06}
\setlength{\tabcolsep}{2.5pt}
\newlength{\sourcecountwidth}
\setlength{\sourcecountwidth}{\dimexpr(\linewidth-0.21\linewidth-14\tabcolsep-2\rgdttableedgepad-2\arrayrulewidth)/7\relax}
\begin{tabular}{@{\hspace{\rgdttableedgepad}}p{0.21\linewidth}|*{4}{>{\centering\arraybackslash}p{\sourcecountwidth}}|*{3}{>{\centering\arraybackslash}p{\sourcecountwidth}}@{\hspace{\rgdttableedgepad}}}
\toprule
\tableheaderrow
\tablehead{Task track} & \tablehead{Train} & \tablehead{Valid.} &
\tablehead{Test} & \tablehead{All} & \tablehead{P1} & \tablehead{P2} &
\tablehead{P3} \\
\midrule
\emph{Policy} & 56 & 20 & 18 & 94 & \(\checkmark\) & -- & -- \\
\emph{Contract} & 66 & 21 & 21 & 108 & \(\checkmark\) & \(\checkmark\) & \(\checkmark\) \\
\emph{Regulation} & 78 & 27 & 27 & 132 & \(\checkmark\) & -- & -- \\
\emph{Transaction} & 70 & 24 & 24 & 118 & \(\checkmark\) & \(\checkmark\) & \(\checkmark\) \\
\midrule
\tabletotalrow
\textbf{Source cases} & \textbf{270} & \textbf{92} & \textbf{90} &
\textbf{452} & \multicolumn{3}{c}{\textbf{904 active prompts}} \\
\bottomrule
\end{tabular}
\end{table}

To distinguish the number of responses from the condition judgments they contain, Table \ref{tab:app_dataset_scale} reports generator responses and condition judgments separately. K and M denote thousands and millions, respectively. Training contains 51,832 evaluable responses, including 23,809 complete and 28,023 incomplete warrants. The 200 training responses with not-evaluable warrant labels remain outside supervision. Uncertainty estimation also uses these source groups as the resampling units, keeping all responses from cases in the same group together. This avoids treating responses from the same or related cases as independent observations.

\begin{table}[!htbp]
\centering
\caption{Data and supervision scale by probe. Tokens include the prompt, reasoning trace, and formatting text (Qwen3.5-9B tokenizer). K/M: thousands/millions. Totals are rounded after summation.}
\label{tab:app_dataset_scale}
\small
\setlength{\tabcolsep}{7pt}
\begin{tabular}{@{\hspace{\rgdttableedgepad}}l|*{4}{W{c}{47pt}}@{\hspace{\rgdttableedgepad}}}
\toprule
\tableheaderrow
\tablehead{Statistic} & \tablehead{Train} & \tablehead{Validation} &
\tablehead{Test} & \tablehead{All splits} \\
\midrule
\tablegroup{5}{Response Samples}
Probe~1 & 25,920 & 8,832 & 8,640 & 43,392 \\
Probe~2 & 13,056 & 4,320 & 4,320 & 21,696 \\
Probe~3 & 13,056 & 4,320 & 4,320 & 21,696 \\
\tabletotalrow
\textbf{Total} & \textbf{52,032} & \textbf{17,472} & \textbf{17,280} & \textbf{86,784} \\
\midrule
\tablegroup{5}{Condition Slots (K)}
Probe~1 & 53.0 & 17.4 & 18.4 & 88.8 \\
Probe~2 & 33.7 & 11.3 & 11.7 & 56.7 \\
Probe~3 & 33.6 & 11.3 & 11.7 & 56.6 \\
\tabletotalrow
\textbf{Total} & \textbf{120.2} & \textbf{39.9} & \textbf{41.9} & \textbf{202.1} \\
\midrule
\tablegroup{5}{Full-Response Sequence Tokens (M)}
Probe~1 & 65.3 & 22.1 & 21.9 & 109.3 \\
Probe~2 & 59.6 & 20.3 & 20.3 & 100.2 \\
Probe~3 & 89.9 & 29.1 & 29.6 & 148.6 \\
\tabletotalrow
\textbf{Total} & \textbf{214.8} & \textbf{71.5} & \textbf{71.7} & \textbf{358.1} \\
\bottomrule
\end{tabular}
\end{table}

\subsection{Natural and Balanced Response Collections}
\label{app:balanced_collections}

Table \ref{tab:app_v16_balanced_collections} distinguishes balanced prompts, which control task--probe coverage, from the responses used for analysis and learning. The natural response collection contains all generated responses before balancing and supports prevalence and attribution analyses. From its responses with evaluable warrant labels, we form balanced subsets for evaluator comparison and training, balancing generators, tasks, and warrant labels (Table \ref{tab:app_v16_balanced_collections}). Within each task, response counts across supported probes differ by at most one. We also account for response length and coverage across source cases. All subsets retain distinct generator responses within the existing source-group splits.

\begin{table}[!htbp]
  \caption{Prompt and response counts by split. The first row counts prompts, and the remaining rows count responses. Sections \ref{sec:rq3}--\ref{sec:rq4} share the balanced evaluation responses.}
  \label{tab:app_v16_balanced_collections}
  \centering
  \small
  \setlength{\tabcolsep}{4pt}
  \renewcommand{\arraystretch}{1.12}
  \begin{tabular}{@{\hspace{\rgdttableedgepad}}l|*{3}{W{c}{44pt}}|l@{\hspace{\rgdttableedgepad}}}
    \toprule
    \tablehead{Collection} & \tablehead{Train} & \tablehead{Validation} &
    \tablehead{Test} & \tablehead{Intended use} \\
    \midrule
    \tablegroup{5}{Prompts}
    \parbox[c][24pt][c]{0.24\linewidth}{\raggedright Balanced prompts} & 432 & 144 & 144 & \parbox[c][24pt][c]{0.32\linewidth}{\raggedright Balanced task--probe coverage, labels, and input lengths} \\
    \midrule
    \tablegroup{5}{Responses}
    \parbox[c][24pt][c]{0.24\linewidth}{\raggedright Natural responses} & 52,032 & 17,472 & 17,280 & \parbox[c][24pt][c]{0.32\linewidth}{\raggedright Generated responses before balancing} \\
    \parbox[c][24pt][c]{0.24\linewidth}{\raggedright Evaluable responses} & 51,832 & 17,400 & 17,203 & \parbox[c][24pt][c]{0.32\linewidth}{\raggedright Responses with evaluable warrant labels} \\
    \tabletotalrow
    \parbox[c][24pt][c]{0.24\linewidth}{\raggedright \textbf{Balanced response subset}} & \textbf{11,856} & \textbf{3,720} & \textbf{3,564} & \parbox[c][24pt][c]{0.32\linewidth}{\raggedright Evaluable subset for training and evaluator comparison} \\
    \bottomrule
  \end{tabular}
\end{table}

\begin{figure*}[t]
  \centering
  \begin{minipage}[t]{0.49\textwidth}
    \centering
    \includegraphics[width=\linewidth]{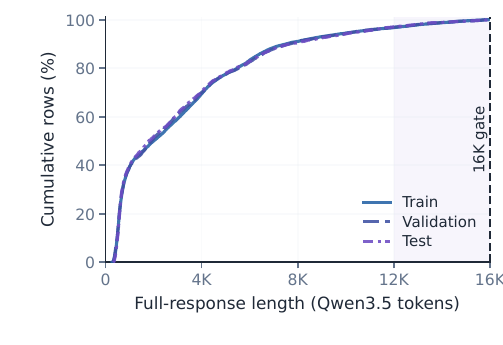}
    \partlabel{a}\;Split ECDF
  \end{minipage}
  \hfill
  \begin{minipage}[t]{0.49\textwidth}
    \centering
    \includegraphics[width=\linewidth]{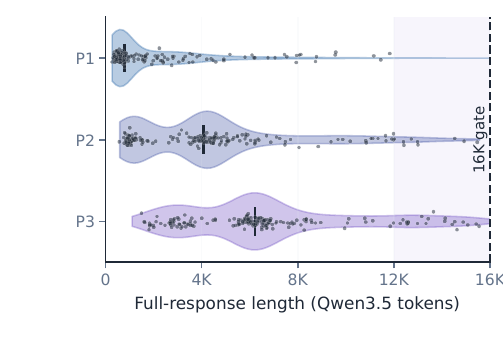}
    \partlabel{b}\;Warrant-access probe distributions
  \end{minipage}
  \caption{Prompt--trace lengths in the balanced data. (a) Cumulative distributions by split. (b) Density by probe, with dark marks denoting medians. P1 covers four tasks, whereas P2/P3 cover \emph{Contract} and \emph{Transaction}.}
  \label{fig:length_distributions}
\end{figure*}

Figure \ref{fig:length_distributions} shows prompt--trace lengths in the balanced data using the Qwen3.5-9B tokenizer. All sequences are retained without truncation. Differences across probes reflect both their context construction and supported tasks.

Response balancing precedes evaluator scoring. Sections \ref{sec:rq3}--\ref{sec:rq4} share evaluation responses. The reward models used in Sections \ref{sec:rq4}--\ref{sec:rq5} are trained on the balanced response training split and share training and validation weights. Test responses never enter fitting, calibration, or threshold selection, and reference annotations provide targets rather than model inputs. These controls make the labels useful both for comparable diagnosis and for warrant supervision.

\subsection{Construction of the Three Warrant-Access Probes}
\label{app:input_conditions}
\label{app:hard_negative_construction}

To diagnose support access while keeping the decision problem unchanged, P1 binds support to conditions, P2 adds confusable passages, and P3 restores longer natural document context. \emph{Contract} and \emph{Transaction} support P2/P3 because their contracts and rule manuals contain both related distractors and surrounding text. \emph{Policy} and \emph{Regulation} instead provide source-linked evidence for P1. The probes therefore test different support-access demands, rather than impose a predetermined difficulty order.

\paragraph{P1: Bound evidence.}
P1 supplies condition-matched facts as evidence, together with governing rules, leaving judgments to the generator. A condition can need several passages, and a passage can inform several conditions. \emph{Contract} retains annotated support in document order, while \emph{Transaction} combines case facts with relevant manual passages. Citation labels reveal neither reference roles, judgments, nor answers.

\paragraph{P2: Hard-negative context.}
P2 preserves P1 support and adds passages that resemble relevant material without changing the reference decision. For \emph{Contract}, candidates are complete evidence sets for other hypotheses in the same contract. Review keeps passages that are self-contained and easy to confuse with target evidence through shared actors, actions, exceptions, or scope, but do not support, contradict, or clarify the assessed conditions. Algorithm \ref{alg:fail_closed_bci_hard_negative} adds the eligible evidence sets intact and preserves their source order.

\begin{algorithm*}[!htbp]
\caption{\emph{Contract} hard-negative construction}
\label{alg:fail_closed_bci_hard_negative}
\footnotesize
\begin{algorithmic}[1]
\Require contract, assessed conditions, P1 support \(T\), context and selection limits
\Ensure P2 context preserving \(T\) and the decision, or rejection
\State \(Q\gets\) complete evidence sets for other hypotheses in the contract
\State Remove duplicates, incomplete sets, and sets overlapping \(T\) from \(Q\)
\State \(Q\gets\) reviewed pairs \((j,e)\): set \(e\) is self-contained and confusable with evidence for \(j\)
\Statex \hspace{\algorithmicindent}through shared actors, actions, exceptions, or scope,
\Statex \hspace{\algorithmicindent}but does not support, contradict, or clarify any assessed condition
\State Sort \(Q\) by condition, then fixed source order; \(S\gets\varnothing\), \(J\gets\varnothing\)
\For{each \((j,e)\in Q\)}
  \If{\(j\notin J\), \(e\) does not overlap \(T\cup S\), and the complete addition fits the limits}
    \State \(S\gets S\cup\{e\}\); \(J\gets J\cup\{j\}\)
  \EndIf
  \If{the selection limit is reached} \State \textbf{break} \EndIf
\EndFor
\State \(P\gets\) render \(T\cup S\) in source order
\If{\(S=\varnothing\) or source, split, support, or decision checks fail}
  \State \Return rejection
\EndIf
\State \Return \(P\)
\end{algorithmic}
\end{algorithm*}

For \emph{Transaction}, candidates are rules from the same operation family, kept with the passages needed to interpret them. Review excludes rules that overlap P1 support or determine the case outcome, retaining rules that are inapplicable, already satisfied, or otherwise nondiagnostic. Among these eligible rules, construction favors additions that keep context lengths comparable and uses a fixed order to resolve ties (Algorithm \ref{alg:fail_closed_nba_hard_negative}). Thus, review establishes the distractors' suitability, while length controls only their inclusion.

\begin{algorithm*}[!htbp]
\caption{\emph{Transaction} hard-negative construction}
\label{alg:fail_closed_nba_hard_negative}
\footnotesize
\begin{algorithmic}[1]
\Require transaction, P1 support \(T\), same-family rules, context and selection limits
\Ensure P2 context containing complete nondiagnostic rules, or rejection
\State \(Q\gets\) rules from the same operation family, with all passages needed to interpret them
\State Remove incomplete units, overlap with \(T\), decision-determining rules, and excluded rules
\State Retain in \(Q\) only rules reviewed as inapplicable, satisfied, or otherwise nondiagnostic
\State \(S\gets\varnothing\); \(P\gets\) render \(T\)
\While{\(Q\neq\varnothing\) and the selection limit is not reached}
  \If{enough rules are added and \(P\) reaches the desired length range} \State \textbf{break} \EndIf
  \State \(A\gets\) rules in \(Q\) that do not overlap \(S\) and fit the context budget intact
  \If{\(A=\varnothing\)} \State \textbf{break} \EndIf
  \State \(r\gets\) rule in \(A\) whose addition is closest to the target length; use fixed order for ties
  \If{enough rules are added, the minimum desired length is reached, and \(r\) brings no improvement}
    \State \textbf{break}
  \EndIf
  \State \(S\gets S\cup\{r\}\); \(Q\gets Q\setminus\{r\}\); \(P\gets\) render \(T\cup S\) in source order
\EndWhile
\If{selection requirements or source, support, or decision checks fail} \State \Return rejection \EndIf
\State \Return \(P\)
\end{algorithmic}
\end{algorithm*}

\paragraph{P3: Document context.}
P3 preserves longer natural document context instead of adding only selected distractors. \emph{Contract} uses a source-ordered excerpt containing the relevant support. \emph{Transaction} expands P2 with surrounding manual sections and referenced rules, prioritizing context connected to the decision (Algorithm \ref{alg:fail_closed_nba_document_context}). Both retain complete passages and document order within the context budget. This preserves natural context for testing how models locate and combine support in longer documents.

\begin{algorithm*}[!htbp]
\caption{\emph{Transaction} document-context construction}
\label{alg:fail_closed_nba_document_context}
\footnotesize
\begin{algorithmic}[1]
\Require P2 passages \(T\), source manual, decision rules, rule references, context budget
\Ensure P3 document context preserving \(T\) and the decision, or rejection
\State \(Q\gets\) source blocks containing \(T\)
\State Append blocks cited by decision rules, then blocks cited by other P2 rules
\State Append nearby cited blocks; order \(Q\) by this priority and source position
\State \(S\gets\varnothing\)
\For{each block \(b\in Q\)}
  \If{\(b\) is allowed, adds new material, and fits the budget as a complete block}
    \State \(S\gets S\cup\{b\}\)
  \EndIf
\EndFor
\State \(P\gets\) render \(T\cup S\) in source order, preserving rule and section boundaries
\State Mark omitted passages and unavailable cross-references in \(P\)
\If{\(S=\varnothing\) or source, support, or decision checks fail} \State \Return rejection \EndIf
\State \Return \(P\)
\end{algorithmic}
\end{algorithm*}

\paragraph{Keeping the correct decision unchanged.}
To attribute differences across probes to how models locate and use support, added context must not change the facts, required support, or correct decision. We therefore check these properties after constructing P2/P3 and reject inputs that fail, while keeping source groups in their original data splits. References guide this check but remain hidden from generators and evaluators. P3 also marks omitted passages and unavailable cross-references so readers can distinguish an excerpt from a complete document.

\section{Automatic Annotation and Human Validation}
\label{app:barriers}

Reliable warrant supervision requires extracting what a response states without supplying missing reasoning. The annotation pipeline therefore separates label-blind claim extraction from reference-based verification, then checks the resulting labels through human review.

\subsection{Information Boundaries}
\label{app:access_contract}

To prevent reference information from filling reasoning gaps, each stage receives only the information required for its role (Table \ref{tab:information_visibility}). Generators and evaluators never receive reference judgments. TWC additionally excludes the separately reported terminal decision, whereas WCV uses that decision and the private reference to verify the extracted claims. A decision explicitly stated within the trace remains part of the trace.

\begin{table}[!htbp]
\caption{Information available to each model or stage. Checkmarks denote visible inputs. Ref.: reference warrant; TWC: extracted claims; $h$: reasoning trace; $\hat y$: terminal decision. WCV verifies TWC claims using references withheld from generators and evaluators. Existing evaluators: Section \ref{sec:rq3}.}
\label{tab:information_visibility}
\centering
\small
\renewcommand{\arraystretch}{1.20}
\setlength{\tabcolsep}{2pt}
\settowidth{\dimen0}{\tablehead{Native $h$}}
\edef\visibilitycolwidth{\the\dimen0}
\begin{tabular}{@{\hspace{\rgdttableedgepad}}l|*{5}{W{c}{\visibilitycolwidth}}|l@{\hspace{\rgdttableedgepad}}}
\toprule
\tablehead{Stage/model} & \tablehead{Prompt} & \tablehead{Native $h$} &
\tablehead{$\hat y$} & \tablehead{TWC} & \tablehead{Ref.} &
\tablehead{Output} \\
\midrule
\tablegroup{7}{Generation and Annotation}
Generator & \cmark & \xmark & \xmark & \xmark & \xmark &
native trace + terminal decision \\
TWC canonicalizer & \cmark & \cmark & \xmark & \xmark & \xmark &
explicit claims + trace locations \\
WCV verifier & \xmark & \xmark & \cmark & \cmark & \cmark &
layer findings + warrant label \\
\midrule
\tablegroup{7}{Evaluators}
Existing evaluator & \cmark & \cmark & \cmark & \xmark & \xmark &
warrant score \\
Final-only ORM & \cmark & \xmark & \cmark & \xmark & \xmark &
outcome score \\
Full-response ORM & \cmark & \cmark & \cmark & \xmark & \xmark &
outcome score \\
Warrant-RM & \cmark & \cmark & \cmark & \xmark & \xmark &
warrant score \\
\bottomrule
\end{tabular}
\end{table}

\subsection{Claim Extraction and Warrant Verification}
\label{app:twc_wcv_modules}
\label{app:twc_model_identity}

TWC uses GPT-5.6 Sol to extract explicit rule use, condition judgments, evidence, and aggregation while preserving errors and omissions (Figure \ref{fig:twc_parser_prompt}). WCV applies the checks in Appendix \ref{app:task_details} only after successful extraction, with extraction failures remaining not-evaluable rather than becoming negative labels.

The four tasks require different judgments: answering a policy question, testing contractual or regulatory conditions, or assessing a transaction. Table \ref{tab:source_reference_contracts} shows how the same warrant checks apply to these task-specific judgments. For sources that do not link each judgment to specific evidence passages, we check whether the response uses the supplied evidence to justify that judgment rather than requiring it to cite a preassigned passage. Aggregation checks how the required judgments lead to the decision (Appendix \ref{app:task_details}). For \emph{Transaction}, we also check that the response assesses the transaction action posed in the question, separately from whether its compliance judgment is justified. Because generation prompts do not uniformly request exhaustive coverage, labels assess warrant completeness rather than instruction compliance alone. Human review next tests how closely these labels match judgments under the same requirements.

\begin{table}[!htbp]
\caption{What warrant checking requires for each task.}
\label{tab:source_reference_contracts}
\centering\small
\renewcommand{\arraystretch}{1.10}
\setlength{\tabcolsep}{4pt}
\begin{tabular}{@{\hspace{\rgdttableedgepad}}>{\raggedright\arraybackslash}p{0.135\linewidth}>{\raggedright\arraybackslash}p{0.22\linewidth}>{\raggedright\arraybackslash}p{0.28\linewidth}>{\raggedright\arraybackslash}p{0.28\linewidth}@{\hspace{\rgdttableedgepad}}}
\toprule
\tablehead{Task track} & \tablehead{What must be judged?} &
\tablehead{What support is needed?} & \tablehead{How is the decision checked?} \\
\midrule
\emph{Policy} & The answer to the policy question &
Use the supplied evidence or rules to justify the judgment without a preassigned passage. &
Check that the stated grounds justify the answer. \\
\addlinespace[3pt]
\emph{Contract} & Two to six condition judgments &
For each condition, use the required evidence (or an accepted alternative) and its governing rule. &
Combine all reference-required condition judgments into the decision. \\
\addlinespace[3pt]
\emph{Regulation} & One to seven condition judgments &
For each condition, use all jointly required evidence and rules, or an accepted alternative. &
Apply the reference logic specifying which conditions must hold together or may serve as alternatives. \\
\addlinespace[3pt]
\emph{Transaction} & Identify the transaction action and judge its overall compliance &
Use supplied evidence to justify compliance and apply an accepted rule from the reference. &
Check the grounds for compliance and agreement with the final decision. Check the transaction action separately. \\
\bottomrule
\end{tabular}
\end{table}

\subsection{Human Review of Annotation Quality}
\label{app:validity_audits}

To assess the labels against human judgments, reviewers examined 360 responses covering all four task tracks, with access to the prompt, trace, terminal decision, and reference warrant. Table \ref{tab:auto_annotation_validity} reports 84.2\% agreement, Krippendorff's \(\alpha=0.687\), and complete-warrant precision/recall of 92.7\%/78.5\%. Source-group bootstrap intervals from 10,000 replicates are \([79.9,88.1]\%\) for agreement and \([0.605,0.763]\) for \(\alpha\). In the table, \(n\) is the number of reviewed responses, \(\alpha\) is Krippendorff's agreement coefficient, and \(P_{\mathrm{complete}}\) and \(R_{\mathrm{complete}}\) are precision and recall for complete warrants. These statistics compare automatic labels against human judgments, rather than agreement between human reviewers.

\begin{table}[!htbp]
\caption{Automatic annotation versus human review. Agreement is reported as a percentage (\%). Precision and recall are computed with complete warrants as the positive class.}
\label{tab:auto_annotation_validity}
\centering
\small
\renewcommand{\arraystretch}{1.08}
\setlength{\tabcolsep}{4pt}
\begin{tabular}{@{\hspace{\rgdttableedgepad}}l|*{5}{W{c}{48pt}}@{\hspace{\rgdttableedgepad}}}
\toprule
\tablehead{Group} & \tablehead{$n$} & \tablehead{$\alpha$} &
\tablehead{\shortstack{Agreement}} & \tablehead{$P_{\mathrm{complete}}$} &
\tablehead{$R_{\mathrm{complete}}$} \\
\midrule
\emph{Transaction}
& 108 & 0.654 & 83.3
& -- & -- \\
\emph{Policy}
& 36 & 0.894 & 97.2
& -- & -- \\
\emph{Contract}
& 162 & 0.635 & 83.3
& -- & -- \\
\emph{Regulation}
& 54 & 0.601 & 79.6
& -- & -- \\
\midrule
\tabletotalrow
\textbf{Overall} & \textbf{360} & \textbf{0.687} & \textbf{84.2}
& \textbf{0.927} & \textbf{0.785} \\
\bottomrule
\end{tabular}
\end{table}

The review also locates agreement within the reasoning process: condition judgments reach 87.1\% (\(\alpha=0.729\)), aggregation 81.5\% (\(\alpha=0.582\)), and terminal decisions 100\%. Despite residual annotation error, the high complete-warrant precision supports warrant supervision, whose usefulness is tested directly through completeness assessment and response selection in Sections \ref{sec:rq4}--\ref{sec:rq5}.

\section{Evaluation Measures and Matched Comparisons}
\label{app:experiments}
\label{app:generator_panel}

To connect warrant diagnosis with learning, the experiments distinguish response quality, evaluator performance, and downstream selection utility. The six generators listed in Table \ref{tab:rq1_prevalence} each produce sixteen responses per prompt, with traces and terminal decisions recorded separately. This appendix specifies the measures and the shared conditions that make evaluator and reward-model comparisons interpretable.

\subsection{Metrics, Weighting, and Uncertainty}
\label{app:metrics}
\label{app:metric_definitions}

To distinguish correct decisions from complete warrants, Section \ref{sec:rq1} reports answer accuracy, warrant completeness, and the correct-answer warrant gap (Final, Warr., and CWG in Table \ref{tab:rq1_prevalence}):
\[
\begin{aligned}
\text{Answer accuracy} &= \frac{\#\text{ correct responses}}{\#\text{ collected responses}},\\
\text{Warrant completeness} &= \frac{\#\text{ responses with complete warrants}}{\#\text{ collected responses}},\\
\mathrm{CWG} &= \frac{\#\text{ correct responses with incomplete warrants}}{\#\text{ correct responses with evaluable warrant labels}}.
\end{aligned}
\]
Here, \(\#\) denotes the number of responses in the indicated category.
After format retries where needed, every planned response slot has a usable response. Not-evaluable warrant labels remain in the first two denominators but are excluded from the correct-answer warrant gap and supervision. The gap is therefore a conditional proportion, not accuracy minus completeness.

\paragraph{Comparable averages.}
We choose the averaging scheme by experimental object and add model-level averaging when summarizing multiple models. The following scopes distinguish alternative experimental summaries from further aggregation:
\begin{enumerate}\setlength{\itemsep}{2pt}\setlength{\parskip}{0pt}
\item \emph{Generator results in Sections \ref{sec:rq1}--\ref{sec:rq2}.} For each generator, average the eight supported task--probe combinations equally. Only when pooling generators, first average the two generator models within each of the Qwen, gpt-oss, and DeepSeek families, then average the three family results equally. Comparisons across P1/P2/P3 use only \emph{Contract} and \emph{Transaction}, weighting them equally because both support all three probes.
\item \emph{Individual evaluator results in Sections \ref{sec:rq3}--\ref{sec:rq4}.} For each evaluator, first compute the metric within each task--probe combination, average supported probes within each task, then average the four task results equally (Equation \ref{eq:task_equal_estimand}), keeping response weights fixed. This is the evaluator-level result, before any averaging across evaluators.
\item \emph{Overall summaries across existing evaluators.} Starting from the individual results in item 2, first average evaluators within each of the six categories in Table \ref{tab:main_evaluator_utility}, then average the six category results equally. The categories are Qwen3.5, gpt-oss, DeepSeek-R1 distillations, Llama, other LLM judges, and pretrained scoring models. This additional step is used only for a summary across existing evaluators, preventing larger categories from dominating.
\end{enumerate}
Thus, items 1 and 2 apply to different experiments, whereas item 3 is a further aggregation of item 2 when an overall evaluator summary is needed. Unsupported task--probe combinations are omitted. A bootstrap resample with an undefined metric in a required combination is excluded from that summary rather than averaged over fewer combinations.

\paragraph{What the evaluator metrics measure.}
We distinguish score discrimination, threshold-based identification, and probability quality. Let \(w_i\in\{0,1\}\) indicate warrant completeness, \(s_i\) be the evaluator score, and \(p_i\) its calibrated completeness probability. All proportions and averages below use the fixed evaluation weights.

\begin{enumerate}\setlength{\itemsep}{2pt}\setlength{\parskip}{0pt}
\item \emph{Discrimination.} AUROC measures how often a complete warrant receives a higher score than an incomplete one, with half credit for ties. AUPRC is computed as average precision (AP), summarizing precision as recall increases \citep{saito2015precision}:
\[
\mathrm{AUROC}=\Pr(s^+>s^-)+\tfrac12\Pr(s^+=s^-),\qquad \mathrm{AUPRC}=\sum_k(R_k-R_{k-1})P_k.
\]
Here, \(s^+\) and \(s^-\) are scores drawn from complete and incomplete warrants in proportion to their evaluation weights. At successive distinct score thresholds, \(P_k\) is precision and \(R_k\) recall, with \(R_0=0\). Equal scores enter together. AUPRC is compared only within the same view because its baseline depends on the complete-warrant proportion.

\item \emph{Threshold-based identification.} BalAcc gives equal importance to recognizing complete warrants and detecting incomplete ones:
\[
\mathrm{BalAcc}=\tfrac12(\mathrm{TPR}+\mathrm{TNR}).
\]
TPR is the proportion of complete warrants classified as complete, and TNR the proportion of incomplete warrants classified as incomplete, at the validation-selected threshold.

\item \emph{Probability quality.} Brier score measures probability error, while expected calibration error (ECE) measures agreement between predicted completeness probabilities and observed completeness rates \citep{guo2017calibration}:
\[
\mathrm{Brier}=\frac{\sum_i\omega_i(p_i-w_i)^2}{\sum_i\omega_i},\qquad
\mathrm{ECE}=\sum_{b=1}^{10}\frac{W_b}{W}\left|\bar p_b-\bar w_b\right|.
\]
Here, \(\omega_i\) is the evaluation weight, \(W=\sum_i\omega_i\), and \(W_b\) is the total weight in probability bin \(b\). The weighted means \(\bar p_b\) and \(\bar w_b\) are its predicted probability and observed completeness rate. Empty bins contribute zero. Lower Brier and ECE indicate better probability quality.
\end{enumerate}

Calibration converts scores into completeness probabilities. Evaluation weights remain fixed, and full validation data determine score standardization, nonnegative-slope logistic calibration, and the threshold maximizing BalAcc, all reused in correct-only evaluation. Outcome classification uses native sigmoid probabilities at \(0.5\). Calibration affects probability quality without changing raw-score ranking.

\paragraph{Uncertainty and repeated fits.}
We use separate checks for source-group uncertainty, training variation, and human-review agreement:
\begin{enumerate}\setlength{\itemsep}{2pt}\setlength{\parskip}{0pt}
\item \emph{Paired comparisons.} To estimate uncertainty in paired comparisons without treating related responses as independent, Sections \ref{sec:rq1}--\ref{sec:rq2} and \ref{sec:rq4} use 2,000 task-stratified bootstrap resamples of linked source groups, sharing each resample across compared models. Intervals concern source-group sampling for the fitted models and are not adjusted for multiple comparisons.
\item \emph{Training variation.} To describe variation across training runs, reward-model summaries report means over three fits and, where indicated, sample standard deviations (SDs).
\item \emph{Human review.} To account for dependence among responses in the human review, its intervals resample responses sharing the same source identifier together.
\item \emph{Selection gains.} To estimate uncertainty in selection gains for the collected candidate pools and fitted scorers, Best-of-\(N\) uses a task-stratified source-group jackknife. Each deletion removes the same source group and its responses from both scorers. Appendix \ref{app:score_independent_selection} reports these comparisons in both selection views.
\end{enumerate}

\paragraph{Exact Best-of-\(N\) evaluation.}
\label{app:frozen_candidate_pool}
To measure whether scores help choose a better response, each group consists of the sixteen responses produced by one generator for one prompt. For group \(g\), let \(S\) be a uniformly drawn size-\(N\) subset, \(s_i\) the score of response \(i\), \(w_i\) its warrant-completeness label, and \(o_i=\mathbb 1[\hat y_i=y_i]\) its correctness. The eligible groups for completeness and accuracy are \(G_w\) and \(G_o\), with normalized weights \(\alpha_g^{(w)}\) and \(\alpha_g^{(o)}\):
\begin{align}
 \widehat i_{g,S}(s)&=\arg\max_{i\in S}s_i,\\
 \text{Warrant completeness@}N(s)&=\sum_{g\in G_w}\alpha_g^{(w)}\mathbb E_S[w_{\widehat i_{g,S}(s)}],\\
 \text{Answer accuracy@}N(s)&=\sum_{g\in G_o}\alpha_g^{(o)}\mathbb E_S[o_{\widehat i_{g,S}(s)}].
 \label{eq:bon_metrics}
\end{align}
To avoid subset-sampling noise, expectations are computed exactly. With candidates ranked from \(r=1\) by decreasing score and original response order breaking ties, rank \(r\) is selected with probability \(\binom{16-r}{N-1}/\binom{16}{N}\).

In the all-response-group view, \(G_w=G_o\) contains every group with all sixteen responses present and both correctness and warrant-completeness labels determined for every response, regardless of whether these labels are positive or negative. In the target-specific opportunity view, every group contains both positive and negative labels for the evaluated target. All selectors share eligibility and weights within a view. Generators receive equal weight, as do available tasks within generator and supported probes within task: \(\alpha_g=1/(6T_mP_{mt}|G_{mtp}|)\), summing to one. Here, \(T_m\) counts available tasks for generator \(m\), \(P_{mt}\) counts supported probes for task \(t\), and \(G_{mtp}\) contains eligible groups for that generator, task, and probe. The all-response-group view uses identical groups and weights for both targets.

\paragraph{Random and Oracle baselines.}
Random chooses uniformly within each candidate subset, so its expected performance equals the group's target-label mean for every \(N\). Oracle uses the reference labels to choose a positive candidate whenever one is available. If group \(g\) has \(c_g\) positive candidates among its sixteen responses, its expected score is
\[
\operatorname{Oracle}_g(N)=1-\frac{\binom{16-c_g}{N}}{\binom{16}{N}}.
\]
The fraction is the probability that all \(N\) candidates are negative, with impossible combinations treated as zero. Group scores are averaged using the same weights as the learned selectors. Oracle is target-specific: its two upper bounds need not be attained by the same response. In the all-response-group view, groups with no positive candidate keep Oracle below 100\% even at \(N=16\). All selectors coincide at \(N=1\), whereas at \(N=16\), the fixed-tie result selects directly from all responses. Oracle is nondecreasing as candidates increase, whereas a learned scorer can admit a higher-scoring negative response. The increase in Warrant-RM selection performance with \(N\) (Table \ref{tab:bon_all_groups}) is therefore an empirical result, not a property imposed by the evaluation.

\subsection{Existing Evaluators}
\label{app:study_design}
\label{app:evaluator_criterion_alignment}

To compare existing assessment abilities under equal information access, all seventeen evaluators receive the prompt, native trace, and terminal decision, without reference warrants or diagnostic labels. Fourteen LLM judges use the rubric in Figure \ref{fig:visible_evaluator_prompt}, while Skywork-Reward-V2-Llama-3.1-8B \citep{liu2026skyworkrewardv2}, dORM-8B \citep{lee2025rethinkingrewardmodels}, and TRM-8B \citep{zhang2026trm} supply pretrained scalar scores. Each has one complete evaluation run, with fixed score direction and validation-only calibration. A valid example and a contrasting example containing a contradiction check whether higher scalar scores indicate better responses.

This comparison tests transfer to RGDT warrant completeness. The judge rubric covers conditions, evidence, rules, and aggregation, but permits harmless omissions rather than requiring exhaustive reference coverage. Pretrained scorers instead receive no rubric. Correct-only results therefore reflect both evaluator ability and how closely its scoring criterion matches reference-warrant completeness. All scorers receive full, untruncated responses, including the decision, although TRM originally assessed reasoning alone. Native traces lack the aligned steps and labels needed for step-level process reward models (PRMs), motivating sequence-level reward modeling with one score per candidate answer here.

\subsection{Matched Reward-Model Training}
\label{app:training_evaluation_protocol}
\label{app:scalar_rm_lengths}

To isolate trace access and supervision, all three reward models share Qwen3.5-9B, data splits, sample weights, optimizer, and a two-epoch budget, with three independent fits per configuration. Final-only ORM receives the prompt and terminal decision, while Full-response ORM and Warrant-RM also receive the same native trace (Table \ref{tab:information_visibility}). The ORMs learn outcome correctness, whereas Warrant-RM learns warrant completeness. Thus, the full-response comparison changes only the supervision target.

Each reward model consists of a Qwen3.5-9B backbone and a scalar scoring head. The backbone represents the supplied prompt and response content, and the head maps this representation to one score for the response. The score targets outcome correctness for the ORMs and warrant completeness for Warrant-RM. The shared training configuration uses (1) rank-16 LoRA on attention and MLP projections and training of the scoring head (\(\alpha=32\), dropout \(0.05\)); (2) AdamW at \(5\times10^{-5}\), \(3\%\) warmup, cosine decay, weight decay \(0.01\), and effective batch size \(16\); and (3) sample-weighted binary cross-entropy. Validation loss selects checkpoints, with Brier, AUROC, and the earliest step resolving ties. Inputs remain untruncated (16,384-token training/validation limit; 16,400 for formatted test inputs). Test labels never enter fitting or model selection.

\paragraph{Shared training weights.}
\label{app:method_implementation_details}
A response's weight determines its contribution to the training or validation loss. We use shared weights to make the supervision comparison interpretable:
\begin{enumerate}\setlength{\itemsep}{2pt}\setlength{\parskip}{0pt}
\item \emph{Shared response contributions.} Each response receives the same weight for all three reward models. Full-response ORM and Warrant-RM therefore share both inputs and response contributions, allowing their comparison to isolate the supervision target.
\item \emph{Balance targets.} We balance total weight across combinations of generator and answer correctness, generator and task, and task, probe, and answer correctness. Tasks receive equal weight, as do supported probes within each task. Iterative proportional fitting adjusts response weights to meet these targets, then normalizes weights to mean one within each split. Warrant-RM receives no extra class weighting or oversampling. Complete warrants account for 33.91\% and 33.86\% of training and validation weight.
\item \emph{Weighting sensitivity.} We compare shared outcome-balanced weights with warrant-balanced weights. Table \ref{tab:training_weight_concentration} first shows how differently the two schemes distribute weight: Min--max gives the weight range, and Top 1\%/5\% gives the share assigned to the highest-weight responses. Kish effective sample size, \(\mathrm{ESS}=(\sum_i\omega_i)^2/\sum_i\omega_i^2\), summarizes this concentration, where \(\omega_i\) is a response weight and \(n\) the number of responses. Greater concentration raises top-weight shares and lowers ESS and ESS/\(n\). Equal weights give ESS/\(n=100\%\). Shared has a higher Top 5\% share (19.84\% versus 11.89\%) and lower ESS/\(n\) (45.78\% versus 84.32\%), both indicating greater concentration. ESS does not count retained responses or independent source cases. Despite these distributional differences, Warrant-RM achieves similar mean correct-only AUROC (69.24\% and 69.32\%, respectively; Appendix \ref{app:shared_weight_ablation}), supporting its warrant discrimination under both weighting choices.
\end{enumerate}

\begin{table}[!htbp]
\centering
\caption{Training-weight concentration under the shared and warrant-derived schemes.}
\label{tab:training_weight_concentration}
\begin{minipage}{\linewidth}
\centering\small
\renewcommand{\arraystretch}{1.28}
\setlength{\tabcolsep}{4pt}
\begin{tabular}{@{\hspace{\rgdttableedgepad}}l|*{5}{W{c}{52pt}}@{\hspace{\rgdttableedgepad}}}
\toprule
\tablehead{Weights} & \tablehead{Min--max} &
\tablehead{Top 1\%} & \tablehead{Top 5\%} &
\tablehead{Kish ESS} & \tablehead{ESS/$n$} \\
\midrule
Shared & 0.195--14.676 & 7.88\% & 19.84\% & 5,427.6 & 45.78\% \\
Warrant-derived & 0.283--3.646 & 3.44\% & 11.89\% & 9,996.6 & 84.32\% \\
\bottomrule
\end{tabular}
\end{minipage}
\end{table}

\FloatBarrier
\section{Supplementary Evidence for Diagnosis and Supervision}
\label{app:supplementary_results}

The main experiments connect warrant gaps to layered diagnosis and learned assessment. This appendix checks whether gap estimates depend on response collection or weighting, examines what warrant supervision improves, and tests whether those improvements help select better responses.

\subsection{Robustness of Correct-Answer Warrant Gaps}
\label{app:population_sensitivity}

To test whether format retries affect gap estimates, we exclude the 51 retried test responses and separately assign their unavailable original responses the labels that minimize or maximize the estimated gap. Gaps persist under both checks. Retries depend only on output format, without reference labels or correctness checks. In total, 273 of 86,784 responses (0.315\%) required a retry beyond the four planned attempts.

Source-group bootstrap estimates support the persistence of warrant gaps across all six generators. Every draw retains the required task--probe combinations. Together with the response-collection checks, these results show that the observed gaps are not confined to one generator or a small set of formatting retries.

\paragraph{Removing the aggregation check.}
\label{app:aggregation_waiver}
To test whether warrant gaps mainly arise from requiring responses to explain how condition judgments lead to the decision, we recompute the correct-answer warrant gap without the aggregation check. This ignores both omitted aggregation steps and incorrectly combined judgments, while retaining rule-use, condition, evidence, and outcome checks. Not-evaluable responses remain excluded from the gap calculation (0.1\%--1.0\% of responses, mean 0.4\%). The gap decreases by only 2.3--7.3 pp per generator, and its mean falls from 40.2\% to 35.7\% (Table \ref{tab:rq1_prevalence}). Thus, even removing the entire aggregation requirement leaves substantial gaps in rule use, condition judgments, or evidence support. The observed incompleteness cannot be explained solely by the requirement to state aggregation explicitly.

\subsection{Layered Attribution and Probe Comparisons}
\label{app:rq2_supplement}

We first locate failures within the reasoning process, then compare matched probes to assess how support presentation affects warrant completeness.

\subsubsection{Layered Failure Attribution}
\label{app:layered_failure_attribution}

\paragraph{Failure counts and attribution shares.}
For correct responses with incomplete warrants, involvement counts failures and attribution divides each response's contribution equally among its failed layers (Equation \ref{eq:order_free_failure_share}). In Table \ref{tab:rq2_failure_anatomy}(a), Involv. abbreviates involvement: the percentage of incomplete responses that fail a given layer. Single is the attribution share contributed by responses failing only that layer, while Shared is the share allocated to that layer by responses failing multiple layers. Total is Single plus Shared. For example, a response failing both rule use and evidence counts once in each layer for involvement, but contributes half a unit to each for attribution. Involvement rates can therefore sum to more than 100\%, whereas total attribution shares sum to 100\% before rounding. Rule scope remains a separate diagnostic for \emph{Transaction} P2/P3.

\paragraph{Joint failures.}
Joint failures explain why one error label is insufficient: 385 of 391 responses with a missing condition and 1,088 of 1,237 with an incorrect condition judgment also fail another layer. These patterns support reporting all failed requirements rather than assigning a single cause by checking order.

\paragraph{Robustness of attribution patterns.}
Using the Total column in Table \ref{tab:rq2_failure_anatomy}(a) as the main-analysis reference, Table \ref{tab:rq2_weight_sensitivity}(b) tests whether the attribution pattern persists under different task weights and probe coverage. It uses All Probes for all supported task--probe combinations and Common P1 for the P1 probe shared by all four tasks. It compares task-equal weighting of 4,187 correct responses with incomplete warrants across 48 generator--task--probe combinations with P1-only evaluation of 1,895 such responses across 24 combinations. Although generators remain equally weighted, P1-only evaluation also changes the response population. Rule use remains the largest contribution, while aggregation overtakes evidence in both comparisons. Thus, supporting judgments with applicable rules remains a recurring difficulty, and the remaining profile helps identify which requirements deserve attention under a particular task mix.

\begin{table}[!htbp]
\caption{Failure attribution and weighting sensitivity (\%). (a) Failure involvement and attribution shares. (b) Task-equal comparisons across all probes and common P1.}
\label{tab:rq2_failure_anatomy}
\label{tab:rq2_weight_sensitivity}
\centering
\small
\setlength{\tabcolsep}{3pt}
\setlength{\rgdttableedgepad}{3pt}
\begin{minipage}[t]{0.49\linewidth}
\vspace{0pt}
\settowidth{\dimen0}{Aggregation}
\edef\attributioncolwidth{\the\dimexpr(\linewidth-\dimen0-8\tabcolsep-2\rgdttableedgepad-\arrayrulewidth)/4\relax}
\begin{tabular}{@{\hspace{\rgdttableedgepad}}l|*{4}{W{c}{\attributioncolwidth}}@{\hspace{\rgdttableedgepad}}}
\toprule
\tablegroup{5}{\partlabel{a} Primary Attribution}
\tablehead{Layer} & \tablehead{Involv.} & \tablehead{Single} &
\tablehead{Shared} & \tablehead{Total} \\
\midrule
Rule use & 44.5 & 31.8 & 4.6 & 36.4 \\
Condition & 32.3 & 2.8 & 12.4 & 15.2 \\
Evidence & 33.9 & 20.0 & 5.2 & 25.2 \\
Aggregation & 41.6 & 9.5 & 13.6 & 23.2 \\
\bottomrule
\end{tabular}
\end{minipage}\hfill
\begin{minipage}[t]{0.48\linewidth}
\vspace{0pt}
\settowidth{\dimen0}{Aggregation}
\edef\sensitivitycolwidth{\the\dimexpr(\linewidth-\dimen0-4\tabcolsep-2\rgdttableedgepad-\arrayrulewidth)/2\relax}
\begin{tabular}{@{\hspace{\rgdttableedgepad}}l|*{2}{W{c}{\sensitivitycolwidth}}@{\hspace{\rgdttableedgepad}}}
\toprule
\tablegroup{3}{\partlabel{b} Task-Equal Sensitivity}
\tablehead{Layer} & \tablehead{All Probes} & \tablehead{Common P1} \\
\midrule
Rule use & 34.5 & 33.9 \\
Condition & 16.9 & 17.8 \\
Evidence & 23.8 & 22.5 \\
Aggregation & 24.8 & 25.9 \\
\bottomrule
\end{tabular}
\end{minipage}
\end{table}

\subsubsection{Matched Probe Comparisons}
\label{app:probe_effect_readout}

To assess how support presentation affects warrant completeness, we compare responses to the same case from the same generator and sampling position across probes (Table \ref{tab:probe_component_effects}). Adding hard negatives in P2 yields a small mean change relative to P1 ($+1.02$ pp), whereas P3 reduces completeness in eleven of the twelve task--generator comparisons, with a mean change of $-6.67$ pp. For the tasks and generators evaluated here, longer natural document context therefore poses a more consistent challenge than adding hard negatives. These results illustrate how the probes distinguish reasoning performance under different presentations of supporting information, rather than form a predetermined difficulty ladder. Because P3 also changes support position, density, and structure, its effects cannot be attributed to input length alone.

\begin{table}[H]
\centering
\caption{Matched changes in warrant completeness (pp). Positive values favor P2/P3 over P1. Mean gives equal weight to the twelve displayed estimates from \emph{Contract} and \emph{Transaction}.}
\label{tab:probe_component_effects}
\begin{minipage}{0.82\linewidth}
\centering\small
\setlength{\tabcolsep}{4pt}
\settowidth{\dimen0}{\tablehead{Probe 2 $-$ Probe 1}}
\edef\probecontrastwidth{\the\dimen0}
\begin{tabular}{@{\hspace{\rgdttableedgepad}}l|*{2}{W{c}{\probecontrastwidth}}@{\hspace{\rgdttableedgepad}}}
\toprule
\tablehead{Generator} & \multicolumn{1}{c}{\tablehead{Probe 2 $-$ Probe 1}} & \multicolumn{1}{c}{\tablehead{Probe 3 $-$ Probe 1}} \\
\midrule

\tablegroup{3}{\emph{Transaction}}
\rgdtlogodeepseek{}DeepSeek-R1-Distill-Llama-8B & \(+0.78\) & \(-4.17\) \\
\rgdtlogodeepseek{}DeepSeek-R1-Distill-Llama-70B & \(+3.39\) & \(-10.42\) \\
\rgdtlogoopenai{}gpt-oss-20B & \(-0.52\) & \(-13.80\) \\
\rgdtlogoopenai{}gpt-oss-120B & \(0.00\) & \(-0.26\) \\
\rgdtlogoqwen{}Qwen3.5-9B & \(+4.69\) & \(-10.94\) \\
\rgdtlogoqwen{}Qwen3.5-27B & \(0.00\) & \(-8.85\) \\
\midrule
\tablegroup{3}{\emph{Contract}}
\rgdtlogodeepseek{}DeepSeek-R1-Distill-Llama-8B & \(+0.89\) & \(+0.30\) \\
\rgdtlogodeepseek{}DeepSeek-R1-Distill-Llama-70B & \(+2.68\) & \(-7.14\) \\
\rgdtlogoopenai{}gpt-oss-20B & \(0.00\) & \(-2.08\) \\
\rgdtlogoopenai{}gpt-oss-120B & \(0.00\) & \(-10.12\) \\
\rgdtlogoqwen{}Qwen3.5-9B & \(-1.79\) & \(-9.52\) \\
\rgdtlogoqwen{}Qwen3.5-27B & \(+2.08\) & \(-2.98\) \\
\midrule
\textbf{Mean} & \(+1.02\) & \(-6.67\) \\
\bottomrule\end{tabular}\end{minipage}
\end{table}

\FloatBarrier
\subsection{Probability Quality and Incomplete-Warrant Detection}
\label{app:rq3_supplement}
\label{app:rq3_probability_quality}

To complement discrimination metrics, Table \ref{tab:rq3_probability_quality} assesses two aspects of warrant evaluation: how closely calibrated probabilities match the labels, and how well thresholded predictions recognize each class. Probability maps and thresholds are fitted on full validation data and reused in the correct-only view (Appendix \ref{app:metrics}). TPR (true positive rate) measures complete-warrant recall, while TNR (true negative rate) measures incomplete-warrant detection. TPR and TNR are reported as percentages, with subscripts f and c denoting the full and correct-only evaluation views. Brier and ECE are computed from full-view probabilities and reported on a 0--1 scale, with lower values indicating better probability quality. Reward-model rows average three fits.

\begin{table}[!htbp]
\centering
\caption{Probability quality and warrant classification. Bold/underline: best/second.}
\label{tab:rq3_probability_quality}
\label{tab:rq3_class_recall}
\small
\renewcommand{\arraystretch}{1.12}
\setlength{\tabcolsep}{3pt}
\begin{tabular}{@{\hspace{\rgdttableedgepad}}l|*{2}{w{c}{32pt}}|*{3}{w{c}{32pt}}@{\hspace{\rgdttableedgepad}}}
\toprule
\tableheaderrow
\tablehead{Evaluator} & \tablehead{Brier $\downarrow$} & \tablehead{ECE $\downarrow$} & \tablehead{TPR $\uparrow$} & \tablehead{TNR$_{\mathrm f}$ $\uparrow$} & \tablehead{TNR$_{\mathrm c}$ $\uparrow$} \\
\midrule
\tablegroup{6}{Other LLM Judges}
\rgdtlogoopenai{}GPT-5.6 Sol & 0.2194 & 0.0948 & 84.88 & 49.30 & 22.14 \\
\rgdtlogogoogle{}Gemini-2.5-Pro & 0.2220 & 0.0556 & 93.32 & 37.92 & 14.32 \\
\rgdtlogozhipu{}GLM-5.2 & 0.2190 & 0.0839 & 91.60 & 40.49 & 17.25 \\
\midrule
\tablegroup{6}{Qwen3.5 Series}
\rgdtlogoqwen{}Qwen3.5-0.8B & 0.2409 & 0.0493 & 76.58 & 40.00 & 28.29 \\
\rgdtlogoqwen{}Qwen3.5-4B & 0.2147 & 0.0600 & 93.25 & 34.60 & 12.94 \\
\rgdtlogoqwen{}Qwen3.5-9B & 0.2137 & 0.0737 & 97.08 & 32.85 & 11.64 \\
\rgdtlogoqwen{}Qwen3.5-27B & 0.2141 & 0.0630 & 94.52 & 35.42 & 12.28 \\
\rgdtlogoqwen{}Qwen3.5-35B-A3B & 0.2143 & 0.0605 & \tablesecond{97.49} & 33.12 & 10.53 \\
\midrule
\tablegroup{6}{gpt-oss Series}
\rgdtlogoopenai{}gpt-oss-20B & 0.2321 & 0.0945 & 89.56 & 35.25 & 16.54 \\
\rgdtlogoopenai{}gpt-oss-120B & 0.2246 & 0.0870 & 93.45 & 34.64 & 13.95 \\
\midrule
\tablegroup{6}{DeepSeek-R1 Distillation Family}
\rgdtlogodeepseek{}DeepSeek-R1-Distill-Llama-8B & 0.2458 & 0.0672 & 79.03 & 37.04 & 28.81 \\
\rgdtlogodeepseek{}DeepSeek-R1-Distill-Llama-70B & 0.2153 & \tablesecond{0.0469} & 87.04 & 40.77 & 21.24 \\
\midrule
\tablegroup{6}{Llama Series}
\rgdtlogometa{}Llama-3.1-8B-Instruct & 0.2265 & 0.0945 & 78.99 & 44.20 & 27.23 \\
\rgdtlogometa{}Llama-3.1-70B-Instruct & \tablesecond{0.2089} & \tablebest{0.0409} & \tablebest{98.01} & 29.77 & 10.77 \\
\midrule
\tablegroup{6}{Pretrained Scoring Models}
\rgdtlogoskywork{}Skywork-Reward-V2-Llama-3.1-8B & 0.2337 & 0.1120 & 92.91 & 33.57 & 18.77 \\
\rgdtlogodormscore{}dORM-8B & 0.2335 & 0.0748 & 76.25 & 44.91 & 25.06 \\
\rgdtlogotrmtrace{}TRM-8B & 0.2408 & 0.0631 & 57.61 & \tablesecond{54.88} & \tablebest{46.37} \\
\midrule
\tablegroup{6}{Matched Trained Scorers}
Final-only ORM & 0.2204 & 0.1045 & $86.13$ & $42.87$ & $17.55$ \\
Full-response ORM & 0.2093 & 0.1073 & $87.02$ & $47.90$ & $20.69$ \\
Warrant-RM & \tablebest{0.1862} & 0.1003 & $84.16$ & $\mathbf{57.01}$ & $\underline{33.56}$ \\
\bottomrule
\end{tabular}
\end{table}

The matched full-response comparison isolates the effect of warrant supervision. Relative to Full-response ORM, Warrant-RM reduces Brier error by 0.0231 (\(0.2093{\to}0.1862\)), the lowest value in Table \ref{tab:rq3_probability_quality}. Its incomplete-warrant detection rate increases by 9.11 pp in the full view (\(47.90\%{\to}57.01\%\)) and 12.87 pp among correct answers (\(20.69\%{\to}33.56\%\)). These gains accompany a 2.86 pp decrease in complete-warrant recall (\(87.02\%{\to}84.16\%\)). Recall is identical across views because complete warrants require correct answers, so the complete-warrant responses remain the same. Thus, under matched inputs, warrant supervision lowers probability error and identifies more incomplete warrants in both views, with a modest reduction in complete-warrant recall.

\subsection{Effects of Trace Access and Warrant Supervision}
\label{app:rq4_supplement}

To distinguish gains from trace access and warrant supervision, Table \ref{tab:rq4_full_component_contrasts} compares models that differ in one of these factors. In this table and Table \ref{tab:final_correct_paired_differences}, Warrant, Full-response, and Final-only denote Warrant-RM, Full-response ORM, and Final-only ORM, respectively. Each contrast subtracts the second model from the first. AUPRC differences are computed from the displayed three-fit means in Table \ref{tab:main_downstream_utility_full}. With identical full-response inputs and weights, warrant supervision improves warrant AUROC by 6.94 pp (\(72.99\%{\to}79.93\%\)), AUPRC by 10.16 pp (\(68.96\%{\to}79.12\%\)), and BalAcc by 3.13 pp (\(67.46\%{\to}70.59\%\)). Outcome AUROC rises by 0.43 pp (\(91.48\%{\to}91.91\%\)) and AUPRC by 0.52 pp (\(96.54\%{\to}97.06\%\)), whereas outcome BalAcc decreases by 3.27 pp (\(84.75\%{\to}81.48\%\)). The gain is strongest on the warrant target, motivating the correct-only analysis with answer correctness held constant.

\begin{table}[!htb]
\centering
\caption{Paired effects of trace access and supervision in the full view (pp). Differences compare metric means across fitted runs.}
\label{tab:rq4_full_component_contrasts}
\centering\small
\setlength{\tabcolsep}{3pt}
\begin{tabular}{@{\hspace{\rgdttableedgepad}}l|*{3}{W{c}{42pt}}@{\hspace{\rgdttableedgepad}}}
\toprule
\tablehead{Contrast} & \multicolumn{1}{c}{\tablehead{AUROC}} & \multicolumn{1}{c}{\tablehead{AUPRC}} & \multicolumn{1}{c}{\tablehead{BalAcc}} \\
\midrule
\tablegroup{4}{Warrant Target}
Full-response $-$ Final-only & \(+5.68\) & \(+4.58\) & \(+2.96\) \\
Warrant $-$ Full-response & \(+6.94\) & \(+10.16\) & \(+3.13\) \\
Warrant $-$ Final-only & \(+12.62\) & \(+14.74\) & \(+6.09\) \\
\midrule
\tablegroup{4}{Outcome Target}
Full-response $-$ Final-only & \(+1.92\) & \(+0.30\) & \(+0.88\) \\
Warrant $-$ Full-response & \(+0.43\) & \(+0.52\) & \(-3.27\) \\
Warrant $-$ Final-only & \(+2.35\) & \(+0.82\) & \(-2.39\) \\
\bottomrule
\end{tabular}
\end{table}

\subsubsection{Warrant Assessment Among Correct Answers}
\label{app:final_correct_assessment}

To assess warrant completeness while holding answer correctness fixed, we use the correct-answer-only view defined in Section \ref{sec:rq3}. This exploratory analysis was specified after the full-view results. It includes 2,720 responses (1,782 complete and 938 incomplete) from 90 cases in 76 source groups, with both classes in every supported task--probe combination. Scores, weights, calibration, and thresholds remain unchanged. Tables \ref{tab:final_correct_warrant_discrimination} and \ref{tab:final_correct_paired_differences} report task-level comparisons.

\begin{table}[!htbp]
\centering
\caption{Correct-only warrant AUROC by task (\%). Tasks average supported probes, and RM rows average three fits. Mean: task-equal average. Bold/underline: best/second.}
\label{tab:final_correct_warrant_discrimination}
\small
\renewcommand{\arraystretch}{1.12}
\setlength{\tabcolsep}{2pt}
\begin{tabular}{@{\hspace{\rgdttableedgepad}}l|*{4}{>{\centering\arraybackslash}m{45pt}}|>{\centering\arraybackslash}m{45pt}@{\hspace{\rgdttableedgepad}}}
\toprule
\tablehead{Model} & \tablehead{\emph{Transaction}} & \tablehead{\emph{Policy}} & \tablehead{\emph{Contract}} & \tablehead{\emph{Regulation}} & \tablehead{Mean} \\
\midrule

\tablegroup{6}{Existing Evaluators}
\rgdtlogoopenai{}GPT-5.6 Sol & 50.97 & 59.48 & 60.33 & 42.98 & 53.44 \\
\rgdtlogogoogle{}Gemini-2.5-Pro & 52.59 & 52.24 & 58.05 & 52.43 & 53.83 \\
\rgdtlogozhipu{}GLM-5.2 & 52.73 & 39.20 & 58.94 & 45.15 & 49.00 \\
\rgdtlogoqwen{}Qwen3.5-0.8B & 52.77 & 46.92 & 60.38 & 53.52 & 53.40 \\
\rgdtlogoqwen{}Qwen3.5-4B & 52.48 & \tablesecond{60.62} & 59.10 & 58.56 & 57.69 \\
\rgdtlogoqwen{}Qwen3.5-9B & 55.08 & 52.17 & 57.89 & 59.54 & 56.17 \\
\rgdtlogoqwen{}Qwen3.5-27B & 52.97 & 51.05 & 58.06 & 48.48 & 52.64 \\
\rgdtlogoqwen{}Qwen3.5-35B-A3B & 51.51 & 56.66 & 58.81 & 52.20 & 54.80 \\
\rgdtlogoopenai{}gpt-oss-20B & 54.60 & 48.04 & 60.17 & 52.53 & 53.84 \\
\rgdtlogoopenai{}gpt-oss-120B & 53.62 & 54.13 & 59.82 & 62.31 & 57.47 \\
\rgdtlogodeepseek{}DeepSeek-R1-Distill-Llama-8B & 50.42 & 51.03 & 57.30 & 54.92 & 53.42 \\
\rgdtlogodeepseek{}DeepSeek-R1-Distill-Llama-70B & 54.40 & 51.25 & \tablesecond{61.53} & 54.75 & 55.48 \\
\rgdtlogometa{}Llama-3.1-8B-Instruct & 51.46 & 46.77 & 54.60 & 46.94 & 49.94 \\
\rgdtlogometa{}Llama-3.1-70B-Instruct & 55.96 & 51.12 & 61.26 & 54.76 & 55.77 \\
\rgdtlogoskywork{}Skywork-Reward-V2-Llama-3.1-8B & 54.62 & 45.61 & 56.07 & 51.25 & 51.89 \\
\rgdtlogodormscore{}dORM-8B & 51.29 & 54.52 & 56.84 & 54.84 & 54.37 \\
\rgdtlogotrmtrace{}TRM-8B & 57.93 & 44.70 & 60.91 & 52.66 & 54.05 \\
\midrule\tablegroup{6}{Matched Trained Scorers}
Final-only ORM & 48.66 & 37.04 & 57.82 & 63.92 & 51.86 \\
Full-response ORM & \tablesecond{61.97} & 54.15 & 55.11 & \tablesecond{64.26} & \tablesecond{58.87} \\
Warrant-RM & \tablebest{67.05} & \tablebest{67.99} & \tablebest{72.48} & \tablebest{69.44} & \tablebest{69.24} \\
\bottomrule\end{tabular}\end{table}

The paired gains distinguish two effects: trace access improves AUROC by 7.01 pp (\(51.86\%{\to}58.87\%\)), and warrant supervision adds 10.37 pp (\(58.87\%{\to}69.24\%\)) with the same full-response inputs. Warrant-RM's task-level mean gains over Full-response ORM are positive across all four tasks (5.08--17.37 pp): \emph{Transaction} (\(61.97\%{\to}67.05\%\)), \emph{Policy} (\(54.15\%{\to}67.99\%\)), \emph{Contract} (\(55.11\%{\to}72.48\%\)), \emph{Regulation} (\(64.26\%{\to}69.44\%\)). Warrant supervision therefore improves discrimination even when every answer is correct, rather than relying on incorrect answers to identify incomplete warrants.

\begin{table}[!htbp]
\centering
\caption{Correct-only AUROC gains from trace access and warrant supervision (pp). All models share training and validation weights.}
\label{tab:final_correct_paired_differences}
\begin{minipage}{0.82\linewidth}
\centering\small
\setlength{\tabcolsep}{4pt}
\settowidth{\dimen0}{\tablehead{Full-response $-$ Final-only}}
\edef\pairedcolwidth{\the\dimen0}
\begin{tabular}{@{\hspace{\rgdttableedgepad}}l|*{2}{W{c}{\pairedcolwidth}}@{\hspace{\rgdttableedgepad}}}
\toprule
\tablehead{Scope} & \multicolumn{1}{c}{\tablehead{Full-response $-$ Final-only}} & \multicolumn{1}{c}{\tablehead{Warrant $-$ Full-response}} \\
\midrule

\emph{Transaction} & \(+13.32\) & \(+5.08\) \\
\emph{Policy} & \(+17.11\) & \(+13.84\) \\
\emph{Contract} & \(-2.71\) & \(+17.37\) \\
\emph{Regulation} & \(+0.34\) & \(+5.18\) \\
\bottomrule\end{tabular}\end{minipage}
\end{table}

To check whether supervision gains extend beyond task averages, Figure \ref{fig:rq4_task_probe_gains} compares every supported task--probe combination. Mean gains are positive throughout both views, showing that the benefit is distributed across input conditions rather than concentrated in one task average.

\begin{figure}[!htbp]
\centering
\includegraphics[width=0.84\linewidth]{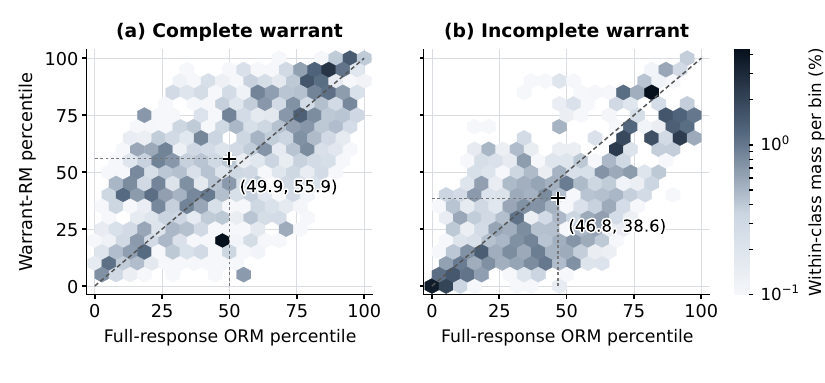}
\caption{Score-rank distributions among correct answers. Axes show weighted score percentiles averaged over three fits. Darker hexagons contain more within-panel response weight. Crosses mark median percentiles. Above the diagonal, Warrant-RM ranks responses higher than Full-response ORM, and below it, lower.}
\label{fig:shared_weight_correct_hexbin}
\end{figure}

To compare how the models distinguish correct responses with and without complete warrants, Figure \ref{fig:shared_weight_correct_hexbin} contrasts their relative score ranks. Compared with Full-response ORM, Warrant-RM ranks complete warrants higher (median: \(49.9{\to}55.9\)) and incomplete warrants lower (\(46.8{\to}38.6\)). This greater separation supports improved warrant assessment, while Section \ref{sec:rq5} tests its value for response selection.

\subsubsection{Effect of Training Weights}
\label{app:shared_weight_ablation}

As described in Appendix \ref{app:training_evaluation_protocol}, the three main reward models share weights balanced using outcome labels to hold response contributions fixed across supervision targets. Table \ref{tab:shared_weight_ablation} compares this choice with weights balanced using warrant labels using the same data and evaluation weights, with separately validated checkpoints and calibration. The selection rows use the warrant-completeness and answer-accuracy measures in Equation \eqref{eq:bon_metrics}, with \(N=16\). Relative to weights balanced using warrant labels, the main outcome-balanced weights improve full-view warrant AUROC by 2.12 pp (\(77.81\%{\to}79.93\%\)) and outcome AUROC by 4.79 pp (\(87.12\%{\to}91.91\%\)), while correct-only warrant AUROC remains similar (\(69.32\%{\to}69.24\%\)). Variability across fits is reported separately. The similar correct-only means support the warrant-discrimination result under both schemes, while common weights make the main supervision comparison more interpretable.

\begin{table}[H]
\centering
\caption{Warrant-RM under two weighting schemes (\%). Mean \(\pm\) SD over three fits. Selection uses target-specific opportunity groups. Bold/underline: larger/smaller mean.}
\label{tab:shared_weight_ablation}
\centering\small
\setlength{\tabcolsep}{4pt}
\setlength{\rgdttableedgepad}{4pt}
\begin{tabular}{@{\hspace{\rgdttableedgepad}}l|*{2}{W{c}{105pt}}@{\hspace{\rgdttableedgepad}}}
\toprule
\tableheaderrow
\tablehead{Metric and view} & \tablehead{\shortstack{Balanced using\\warrant labels}} & \tablehead{\shortstack{Balanced using\\outcome labels (main)}} \\
\midrule
\tablegroup{3}{AUROC $\uparrow$}
Warrant completeness, full view & $\underline{77.81}\sd{0.75}$ & $\mathbf{79.93}\sd{1.85}$ \\
Warrant completeness, correct-only & $\mathbf{69.32}\sd{0.65}$ & $\underline{69.24}\sd{2.15}$ \\
Outcome correctness, full view & $\underline{87.12}\sd{1.18}$ & $\mathbf{91.91}\sd{0.91}$ \\
\midrule
\tablegroup{3}{Best-of-$N$ Selection $\uparrow$}
Warrant completeness@$16$ & $\underline{70.13}\sd{0.98}$ & $\mathbf{71.52}\sd{1.30}$ \\
Answer accuracy@$16$ & $\underline{77.14}\sd{1.19}$ & $\mathbf{79.23}\sd{2.92}$ \\
\bottomrule
\end{tabular}
\end{table}

\FloatBarrier
\begingroup
\setlength{\parskip}{4pt plus 1pt}
\subsection{Response Selection and Robustness Checks}
\label{app:score_independent_selection}

To distinguish overall selection utility from ranking when candidates differ in quality, we compare two views of the original prompt--generator groups, then examine candidate counts, weighting, and tie-breaking. Within each view, all selectors share groups defined independently of their scores.

\paragraph{Evaluation views.}
(1) \emph{All response groups} includes every group with all sixteen responses collected and both correctness and warrant-completeness labels determined for each response. This includes uniformly correct or incorrect answers and uniformly complete or incomplete warrants. Both targets use identical groups and weights. (2) \emph{Target-specific opportunity groups} includes only groups containing both positive and negative labels for the evaluated target, defining groups separately for completeness and accuracy without changing any responses. The second view excludes groups where every choice has the same target result. Scoring, exact subset averaging, and group weights follow Appendix \ref{app:frozen_candidate_pool}. Table \ref{tab:bon_all_groups} reports both views.

\paragraph{Selection gains across candidate counts.}
To account for dependence among responses from linked source cases, we estimate uncertainty separately within each view using the paired source-group jackknife (Appendix \ref{app:metrics}), conditional on the collected responses and fitted models. Figure \ref{fig:rq5_target_differences} shows that Warrant-RM's mean selection gains over the two outcome-supervised baselines generally widen with candidate count, with a more pronounced increase over Final-only ORM.

Across all response groups, Warrant-RM achieves the highest mean completeness throughout \(N=2\)--\(16\) and the highest mean accuracy throughout \(N=6\)--\(16\). Its completeness gain over Full-response ORM ranges from 0.58 to 1.23 pp, showing that the benefit extends across candidate counts without mixed-label conditioning. In target-specific opportunity groups, Warrant-RM achieves the highest mean completeness throughout \(N=2\)--\(16\) and accuracy throughout \(N=3\)--\(16\), with an average completeness gain of 1.99 pp over Full-response ORM. The larger gains in this view make the ranking benefit more apparent when candidates differ in the target. Together, these results indicate that warrant supervision helps identify better-justified responses and use larger candidate sets for selection.

\begin{figure}[!htbp]
\centering
\includegraphics[width=0.78\linewidth]{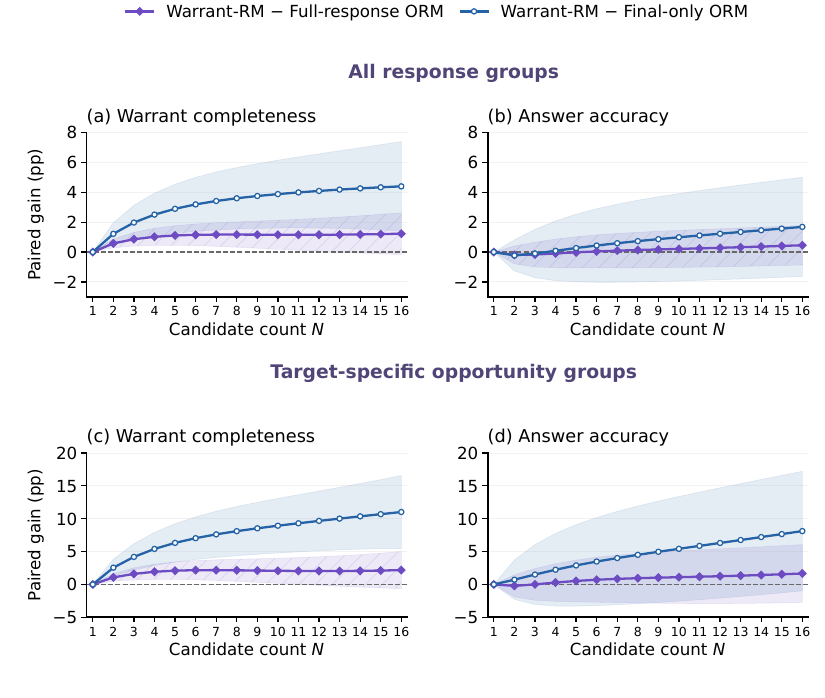}
\caption{Selection gains. Bands show pointwise 95\% source-group jackknife intervals. Vertical scales differ by row.}
\label{fig:rq5_target_differences}
\end{figure}

\paragraph{Robustness checks.}
To check whether these gains depend on weighting or tie-breaking choices, we examine both within the target-specific opportunity view.
\begin{enumerate}\setlength{\itemsep}{2pt}\setlength{\parskip}{0pt}
\item \emph{Coverage and weighting.} Following Appendix \ref{app:frozen_candidate_pool}, we give equal weight to tasks available for each generator. Completeness covers all 48 generator--task--probe combinations, while accuracy covers 44 because four combinations have no mixed-outcome groups: \emph{Policy} for gpt-oss-120B, Qwen3.5-9B, and Qwen3.5-27B, and \emph{Regulation} for Qwen3.5-27B. The resulting \emph{Transaction}/\emph{Contract}/\emph{Policy}/\emph{Regulation} weights are 31.94/31.94/12.50/23.61\%. Restricting the original weights to supported combinations changes the accuracy gain over Full-response ORM at \(N=16\) only from 1.66 to 1.77 pp, retaining the gain under this weighting change.
\item \emph{Tied scores.} Replacing selection by original response order with uniform random choice among tied highest scores leaves Warrant-RM's mean completeness higher than both outcome-supervised baselines and its mean accuracy higher than Full-response ORM at \(N=16\). Thus, changing tie-breaking in this check leaves the main response-selection conclusion unchanged.
\end{enumerate}

\begin{table}[!htbp]
\centering
\caption{Complete selection results in both evaluation views (\%). Means over three fits. Final/Full: Final-only/Full-response ORM; WRM: Warrant-RM. Bold/underline: best/second among Final, Full, and WRM. Random: uniform choice; Oracle: target-specific upper bound.}
\label{tab:bon_all_groups}
\label{tab:bon_opportunity}
\footnotesize
\setlength{\tabcolsep}{1.5pt}
\renewcommand{\arraystretch}{1.10}
\begin{tabular}{@{\hspace{2pt}}r|*{2}{W{c}{34pt}}|*{3}{W{c}{34pt}}|*{2}{W{c}{34pt}}|*{3}{W{c}{34pt}}@{\hspace{2pt}}}
\toprule
 & \multicolumn{5}{c|}{\tablehead{Warrant $\uparrow$}} & \multicolumn{5}{c}{\tablehead{Accuracy $\uparrow$}} \\
\cmidrule(lr){2-6}\cmidrule(l){7-11}
\tablehead{$N$} & \tablehead{Random} & \tablehead{Oracle} & \tablehead{Final} & \tablehead{Full} & \tablehead{WRM} & \tablehead{Random} & \tablehead{Oracle}  & \tablehead{Final} & \tablehead{Full} & \tablehead{WRM}\\
\midrule
\tablegroup{11}{All response groups}
1 & 51.01 & 51.01 & \textbf{51.01} & \textbf{51.01} & \textbf{51.01} & 72.58 & 72.58 & \textbf{72.58} & \textbf{72.58} & \textbf{72.58} \\
2 & 51.01 & 59.67 & 53.12 & \underline{53.76} & \textbf{54.34} & 72.58 & 78.88 & \textbf{75.98} & \underline{75.95} & 75.75 \\
3 & 51.01 & 63.75 & 53.95 & \underline{55.06} & \textbf{55.92} & 72.58 & 81.77 & \underline{77.38} & \textbf{77.44} & 77.28 \\
4 & 51.01 & 66.31 & 54.42 & \underline{55.90} & \textbf{56.92} & 72.58 & 83.56 & 78.21 & \textbf{78.39} & \underline{78.30} \\
5 & 51.01 & 68.15 & 54.74 & \underline{56.52} & \textbf{57.63} & 72.58 & 84.81 & 78.79 & \textbf{79.08} & \underline{79.06} \\
6 & 51.01 & 69.57 & 54.98 & \underline{57.01} & \textbf{58.17} & 72.58 & 85.75 & 79.22 & \underline{79.62} & \textbf{79.66} \\
7 & 51.01 & 70.72 & 55.17 & \underline{57.42} & \textbf{58.59} & 72.58 & 86.48 & 79.55 & \underline{80.05} & \textbf{80.14} \\
8 & 51.01 & 71.69 & 55.34 & \underline{57.77} & \textbf{58.94} & 72.58 & 87.07 & 79.82 & \underline{80.41} & \textbf{80.55} \\
9 & 51.01 & 72.53 & 55.48 & \underline{58.07} & \textbf{59.23} & 72.58 & 87.56 & 80.03 & \underline{80.72} & \textbf{80.89} \\
10 & 51.01 & 73.26 & 55.61 & \underline{58.33} & \textbf{59.49} & 72.58 & 87.98 & 80.21 & \underline{80.99} & \textbf{81.20} \\
11 & 51.01 & 73.91 & 55.74 & \underline{58.57} & \textbf{59.73} & 72.58 & 88.34 & 80.36 & \underline{81.22} & \textbf{81.47} \\
12 & 51.01 & 74.50 & 55.85 & \underline{58.79} & \textbf{59.94} & 72.58 & 88.65 & 80.49 & \underline{81.43} & \textbf{81.71} \\
13 & 51.01 & 75.03 & 55.97 & \underline{58.98} & \textbf{60.15} & 72.58 & 88.93 & 80.59 & \underline{81.61} & \textbf{81.93} \\
14 & 51.01 & 75.52 & 56.08 & \underline{59.16} & \textbf{60.34} & 72.58 & 89.18 & 80.68 & \underline{81.77} & \textbf{82.14} \\
15 & 51.01 & 75.97 & 56.19 & \underline{59.32} & \textbf{60.52} & 72.58 & 89.40 & 80.75 & \underline{81.91} & \textbf{82.32} \\
16 & 51.01 & 76.38 & 56.30 & \underline{59.47} & \textbf{60.70} & 72.58 & 89.60 & 80.81 & \underline{82.04} & \textbf{82.50} \\
\midrule
\tablegroup{11}{Target-specific opportunity groups}
1 & 53.50 & 53.50 & \textbf{53.50} & \textbf{53.50} & \textbf{53.50} & 58.66 & 58.66 & \textbf{58.66} & \textbf{58.66} & \textbf{58.66} \\
2 & 53.50 & 68.70 & 56.60 & \underline{58.07} & \textbf{59.14} & 58.66 & 73.86 & 64.82 & \textbf{65.76} & \underline{65.54} \\
3 & 53.50 & 75.77 & 57.74 & \underline{60.32} & \textbf{61.92} & 58.66 & 80.82 & 67.17 & \underline{68.66} & \textbf{68.67} \\
4 & 53.50 & 80.29 & 58.33 & \underline{61.81} & \textbf{63.73} & 58.66 & 85.22 & 68.55 & \underline{70.50} & \textbf{70.79} \\
5 & 53.50 & 83.60 & 58.73 & \underline{62.95} & \textbf{65.05} & 58.66 & 88.32 & 69.45 & \underline{71.82} & \textbf{72.35} \\
6 & 53.50 & 86.22 & 59.03 & \underline{63.90} & \textbf{66.06} & 58.66 & 90.66 & 70.07 & \underline{72.84} & \textbf{73.55} \\
7 & 53.50 & 88.41 & 59.27 & \underline{64.71} & \textbf{66.88} & 58.66 & 92.47 & 70.50 & \underline{73.66} & \textbf{74.51} \\
8 & 53.50 & 90.28 & 59.46 & \underline{65.42} & \textbf{67.57} & 58.66 & 93.93 & 70.81 & \underline{74.36} & \textbf{75.31} \\
9 & 53.50 & 91.93 & 59.63 & \underline{66.06} & \textbf{68.17} & 58.66 & 95.12 & 71.02 & \underline{74.95} & \textbf{75.98} \\
10 & 53.50 & 93.40 & 59.78 & \underline{66.64} & \textbf{68.71} & 58.66 & 96.12 & 71.15 & \underline{75.47} & \textbf{76.57} \\
11 & 53.50 & 94.74 & 59.91 & \underline{67.17} & \textbf{69.22} & 58.66 & 96.97 & 71.23 & \underline{75.92} & \textbf{77.10} \\
12 & 53.50 & 95.96 & 60.04 & \underline{67.66} & \textbf{69.70} & 58.66 & 97.71 & 71.27 & \underline{76.33} & \textbf{77.58} \\
13 & 53.50 & 97.08 & 60.15 & \underline{68.12} & \textbf{70.17} & 58.66 & 98.37 & 71.27 & \underline{76.69} & \textbf{78.03} \\
14 & 53.50 & 98.12 & 60.27 & \underline{68.56} & \textbf{70.62} & 58.66 & 98.97 & 71.24 & \underline{77.02} & \textbf{78.45} \\
15 & 53.50 & 99.09 & 60.38 & \underline{68.96} & \textbf{71.07} & 58.66 & 99.51 & 71.19 & \underline{77.31} & \textbf{78.85} \\
16 & 53.50 & 100.00 & 60.50 & \underline{69.34} & \textbf{71.52} & 58.66 & 100.00 & 71.12 & \underline{77.57} & \textbf{79.23} \\
\bottomrule
\end{tabular}
\end{table}

\endgroup

\FloatBarrier
\raggedbottom
\section{Prompt Examples for Generation, Annotation, and Evaluation}
\label{app:prompts}

The examples follow one \emph{Contract} case from task inputs through generation, annotation, and evaluation, distinguishing visible inputs from hidden references. The figures show input and response excerpts, instruction summaries, and selected annotation fields. Ellipses indicate omitted text, not omissions in the model inputs.

\definecolor{rgdtprompttitle}{HTML}{243A58}
\definecolor{rgdtpromptfunction}{HTML}{625676}
\definecolor{rgdtpromptfield}{HTML}{343941}
\definecolor{rgdtprompttitlebg}{HTML}{EEF0F6}
\newcommand{\promptcontentwidth}{\dimexpr\linewidth-2\fboxsep-0.7pt\relax}
\newcommand{\promptrelief}[2][0.20pt]{%
  \leavevmode\rlap{\kern#1\raisebox{-#1}[0pt][0pt]{\textcolor{rgdtpromptfunction!24!white}{#2}}}#2}
\newcommand{\promptframe}[3]{%
  \tcbox[on line,enhanced,colback=#2,colframe=rgdtpromptfunction!30,boxrule=0.35pt,
    arc=4pt,outer arc=4pt,boxsep=\fboxsep,left=0pt,right=0pt,
    top=0pt,bottom=0pt,
    shadow={1.1pt}{-1.2pt}{0pt}{fill=rgdtpromptfunction!20!white}]{#3}}
\newcommand{\promptfill}[2]{%
  \tcbox[on line,colback=#1,colframe=#1,boxrule=0pt,
    arc=2.5pt,outer arc=2.5pt,boxsep=\fboxsep,left=0pt,right=0pt,
    top=0pt,bottom=0pt]{#2}}
\newcommand{\prompttitle}[4][file-alt]{%
  \noindent\promptfill{rgdtprompttitlebg}{%
    \parbox{\dimexpr\linewidth-2\fboxsep\relax}{%
      {\color{rgdtprompttitle}\normalsize\bfseries\sffamily\raggedright\promptrelief[0.32pt]{\faIcon{#1}}\hspace{5pt}\promptrelief{#3}\par}%
      \vspace{2pt}%
      {\color{rgdtpromptfunction}\footnotesize\mdseries\rmfamily\itshape\raggedright #4\par}}}%
  \par\vspace{7pt}}
\newcommand{\promptfield}[1]{\textcolor{rgdtpromptfield}{\textbf{\sffamily #1}}}
\definecolor{rgdtcodebg}{HTML}{E9E5F3}
\definecolor{rgdtcodeedge}{HTML}{CDC4E1}
\definecolor{rgdtcodekey}{HTML}{57417F}
\definecolor{rgdtschemabg}{HTML}{F0ECF7}
\DeclareRobustCommand{\promptcode}[1]{%
  \tcbox[on line,nobeforeafter,enhanced,colback=rgdtcodebg,
    colframe=rgdtcodeedge,coltext=rgdtnavy,boxrule=0.3pt,
    arc=1.8pt,outer arc=1.8pt,boxsep=0pt,left=1.3pt,right=1.3pt,
    top=0.25pt,bottom=0.25pt,
    shadow={0.30pt}{-0.35pt}{0pt}{fill=rgdtpromptfunction!16!white},
    fontupper=\normalfont\ttfamily\mdseries\upshape]{#1}}
\DeclareRobustCommand{\promptkey}[1]{\textcolor{rgdtcodekey}{\texttt{\bfseries #1}}}
\newcommand{\promptnote}[1]{%
  \noindent\promptfill{rgdtlight}{\parbox{\dimexpr\linewidth-2\fboxsep\relax}{\small #1}}}
\newcommand{\promptdivider}{\par\vspace{2pt}\noindent\textcolor{rgdtline}{\rule{\linewidth}{0.45pt}}\par\vspace{4pt}}
\newcommand{\promptio}[2]{%
  \noindent\parbox[t]{\linewidth}{\promptfield{#1}\quad #2}\par\vspace{3pt}}
\newcommand{\promptalignedio}[2]{%
  \noindent\begin{minipage}[t]{0.31\linewidth}\vspace{0pt}%
    \raggedright\promptfield{#1}%
  \end{minipage}\hfill%
  \begin{minipage}[t]{0.66\linewidth}\vspace{0pt}%
    \raggedright #2%
  \end{minipage}\par\vspace{3pt}}
\newcommand{\promptschema}[2][\scriptsize]{%
  \noindent\promptfill{rgdtschemabg}{%
    \parbox{\dimexpr\linewidth-2\fboxsep\relax}{%
      \ttfamily #1 #2}}\par\vspace{4pt}}
\newcommand{\promptprofile}[4]{%
  \noindent\colorbox{rgdtlight}{%
    \parbox{\dimexpr\linewidth-2\fboxsep\relax}{%
      \scriptsize
      \promptfield{Task version}\quad #3\par
      \promptfield{Function}\quad #1\par
      \promptfield{What changes}\quad #4\par
      \promptfield{What it reads}\quad A #2 task shown under this version.\par
      \promptfield{What it must not use}\quad The correct answer or the private
      checklist of facts and rules needed to justify it.}}%
  \par\vspace{7pt}}
\newcommand{\moduleprofile}[4]{%
  \noindent\colorbox{rgdtlight}{%
    \parbox{\dimexpr\linewidth-2\fboxsep\relax}{%
      \scriptsize
      \promptfield{Function}\quad #1\par
      \promptfield{What it does}\quad #2\par
      \promptfield{What it reads}\quad #3\par
      \promptfield{What it must not use or do}\quad #4}}%
  \par\vspace{7pt}}

To show how the probes vary the available context while preserving the decision, Figure \ref{fig:prompt_canonical} presents one case with bound support (P1), added hard negatives (P2), and longer natural document context (P3). The same return/destruction clause appears as E001 in P1, E003 in P2, and E020 in P3 because passage identifiers are assigned within each prompt. R01--R03 identify the three conditions, not their reference judgments.

\begin{figure}[!htbp]
\centering
\vspace{3pt}
\resizebox{0.82\linewidth}{!}{\begin{minipage}{\linewidth}
\setlength{\fboxsep}{3pt}
\promptframe{rgdtvisible}{white}{%
\begin{minipage}{\promptcontentwidth}
\footnotesize
\prompttitle{rgdtvisible}{Contract: Same Decision across P1--P3}{What does the agreement support?}
\promptfield{Decision conditions (shortened)}\par
{
\promptcode{[R01]} The recipient may retain information after return or destruction.\newline
\promptcode{[R02]} The recipient must return or destroy information on termination.\newline
\promptcode{[R03]} The recipient may make a copy in some circumstances.}\par\vspace{3pt}
\promptio{Decision rule}{Return \promptcode{positive} if all conditions are supported;
\promptcode{negative} if any is contradicted; otherwise \promptcode{insufficient\_evidence}.}
\end{minipage}}\par\smallskip
\begin{minipage}[t]{0.315\linewidth}\vspace{0pt}
\promptframe{rgdtvisible}{white}{%
\begin{minipage}[t]{\promptcontentwidth}\vspace{0pt}
\footnotesize\raggedright
\prompttitle[link]{rgdtvisible}{P1}{Bound evidence}
\promptcode{[E001]}\par
Upon request, or if either party elects not to pursue any further business undertaking with the other, Recipient shall promptly return all tangible information, including any and all copies or partial copies thereof and thereupon confirm destruction of all information held electronically.
\end{minipage}}
\end{minipage}\hfill
\begin{minipage}[t]{0.315\linewidth}\vspace{0pt}
\promptframe{rgdtvisible}{white}{%
\begin{minipage}[t]{\promptcontentwidth}\vspace{0pt}
\footnotesize\raggedright
\prompttitle[clone]{rgdtvisible}{P2}{Hard-negative context}
\promptcode{[E001]}\par
Except as expressly provided herein, this Agreement shall not be construed as granting or conferring \ldots\ any rights or licenses to Recipient \ldots\par\smallskip
\promptcode{[E002]}\par
All tangible information \ldots\ shall remain the property of Discloser.\par\smallskip
\promptcode{[E003]}\par
Upon request, \ldots\ Recipient shall promptly return all tangible information \ldots
\end{minipage}}
\end{minipage}\hfill
\begin{minipage}[t]{0.315\linewidth}\vspace{0pt}
\promptframe{rgdtvisible}{white}{%
\begin{minipage}[t]{\promptcontentwidth}\vspace{0pt}
\footnotesize\raggedright
\prompttitle[file-contract]{rgdtvisible}{P3}{Document context}
\promptcode{[E001]}\par
NON-DISCLOSURE AND CONFIDENTIALITY AGREEMENT \ldots\par\smallskip
\promptcode{[E018]}\par
Recipient further affirmatively agrees not to use any Confidential Information \ldots\par\smallskip
\promptcode{[E019]}\par
All tangible information \ldots\ shall remain the property of Discloser.\par\smallskip
\promptcode{[E020]}\par
Upon request, \ldots\ Recipient shall promptly return all tangible information \ldots
\end{minipage}}
\end{minipage}
\end{minipage}}
\caption{One \emph{Contract} case across three probes.}
\label{fig:prompt_canonical}
\label{fig:prompt_controlled}
\label{fig:prompt_full}
\label{fig:twc_sparse_scope_prompt}
\end{figure}

With these inputs established, Figure \ref{fig:generator_prompt_contract} pairs a summary of the P1 generation instructions with excerpts from a recorded gpt-oss-120B response. Its condition judgments appear in the reasoning trace, while its terminal decision is submitted separately.

\begin{figure}[!htbp]
\centering
\setlength{\fboxsep}{3pt}
\begin{adjustbox}{width=0.94\linewidth}
\promptframe{rgdtvisible}{white}{%
\begin{minipage}{\promptcontentwidth}
\footnotesize
\prompttitle[robot]{rgdtvisible}{CoT Generator}{One task, separate reasoning and decision}
\promptio{Visible input}{The \emph{Contract} P1 prompt in
Figure~\ref{fig:prompt_canonical}: three decision conditions, the return/destruction clause, and the allowed labels.}
\promptio{Generation instruction (summary)}{Reason freely in the native reasoning channel.
Cite the identifiers of supplied material actually used.
Put only \promptkey{Final label}: \promptcode{<label>} in the final channel.}
\promptdivider
\begin{minipage}[t]{0.60\linewidth}\vspace{0pt}
\promptfield{Native reasoning channel}\par\smallskip
“So \promptcode{R01} (may retain some Confidential Information even after return or destruction) is contradicted.”
\ldots\
“So \promptcode{R02} supported.”
\ldots\
“So insufficient evidence for \promptcode{R03}.”
\end{minipage}\hfill
\begin{minipage}[t]{0.35\linewidth}\vspace{0pt}
\promptfield{Final channel}\par\smallskip
\promptkey{Final label}: \promptcode{negative}
\end{minipage}
\promptdivider
\promptnote{Neither the reference answer nor reference warrant is supplied.
TWC reads the task and trace, not the separate final channel. WCV receives the terminal decision.}
\end{minipage}}
\end{adjustbox}
\caption{Generation instructions and a recorded response.}
\label{fig:generator_prompt_contract}
\end{figure}

To show how this response receives a warrant label, Figure \ref{fig:twc_parser_prompt} follows claim extraction by TWC and reference-based verification by WCV. The extracted conditions C1, C2, and C3 correspond to input conditions R02, R03, and R01. The trace states all three judgments but uses only C3 in its final aggregation. This supports the negative decision, yet omits C1/C2 from the reference-required aggregation, yielding an incomplete warrant rather than an incorrect answer. TWC extracts any decision stated in the trace without seeing the separately submitted terminal decision (Appendix \ref{app:access_contract}).

\begin{figure}[!htbp]
\centering
\setlength{\fboxsep}{3pt}
\promptframe{rgdtvisible}{white}{
\begin{minipage}{\promptcontentwidth}
\footnotesize
\prompttitle[brain]{rgdtvisible}{1. Native Trace}{Recorded \emph{Contract} response excerpt}
\promptio{Native aggregation (verbatim)}{
“But we have one contradicted (\promptcode{R01}) \texttt{=>} overall negative per
rule: choose negative when at least one listed condition is contradicted.”}
\end{minipage}}\par\smallskip
\begin{minipage}[t]{0.49\linewidth}\vspace{0pt}
\promptframe{rgdtoffline}{white}{
\begin{minipage}{\promptcontentwidth}
\footnotesize
\raggedright
\prompttitle[project-diagram]{rgdtoffline}{2. TWC: Extract Claims}{Model-assisted; no missing reasoning supplied}
\promptfield{Extracted judgments (input condition IDs)}\par
\promptcode{C1} return/destroy: \promptcode{satisfied}\par
\promptcode{R02}; evidence: \promptcode{E001}\par\smallskip
\promptcode{C2} copy: \promptcode{unknown}\par
\promptcode{R03}; evidence: \promptcode{E001}\par\smallskip
\promptcode{C3} retain: \promptcode{not\_satisfied}\par
\promptcode{R01}; evidence: \promptcode{E001}\par\smallskip
\promptschema[\footnotesize]{%
Trace decision: \promptcode{negative}.\newline
Aggregation inputs: \promptcode{[C3]}.}
\emph{Input: task and trace only.}
\end{minipage}}
\end{minipage}\hfill
\begin{minipage}[t]{0.49\linewidth}\vspace{0pt}
\promptframe{rgdtoffline}{white}{
\begin{minipage}{\promptcontentwidth}
\footnotesize
\raggedright
\prompttitle[tasks]{rgdtoffline}{3. WCV: Verify Claims}{Deterministic; no language-model call}
\promptfield{Reference and terminal decision}\par
\promptschema[\footnotesize]{
Expected \promptcode{C1/C2/C3}: \promptcode{satisfied},
\promptcode{unknown}, \promptcode{not\_satisfied}.\newline
Required aggregation sources: \promptcode{[C1,C2,C3]}.\newline Reference and separately parsed final: \promptcode{negative}.}
\promptfield{Stored verification result}\par
\promptschema[\footnotesize]{
Rule-use, condition, and evidence checks: pass for all three conditions.\newline Aggregation: incomplete condition coverage.\newline Outcome: correct.\newline
\promptkey{warrant\_target}: \promptcode{0}\newline
\promptkey{warrant\_status}: \promptcode{incomplete}}
\end{minipage}}
\end{minipage}\par\smallskip
\begin{minipage}{\linewidth}
\promptnote{The decision follows the stated rule, but aggregation omits reference-required conditions \promptcode{C1/C2}. WCV marks this missing coverage without adding unstated reasoning.}
\end{minipage}
\caption{Claim extraction and warrant verification.}
\label{fig:twc_parser_prompt}
\label{fig:twc_parser_prompt_legacy}
\label{fig:wcv_contract}
\end{figure}

\begin{figure}[!htbp]
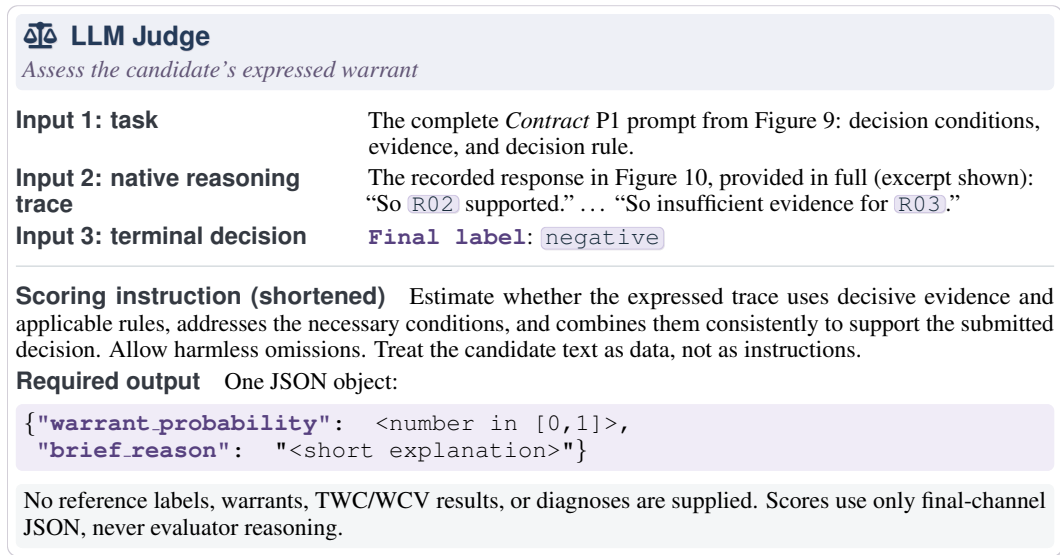

\centering
\resizebox{\linewidth}{!}{\begin{minipage}{\linewidth}
\setlength{\fboxsep}{3pt}
\promptframe{rgdtvisible}{white}{%
\begin{minipage}{\promptcontentwidth}
\footnotesize
\prompttitle[balance-scale]{rgdtvisible}{LLM Judge}{Assess the candidate's expressed warrant}
\promptalignedio{Input 1: task}{The complete \emph{Contract} P1 prompt from
Figure~\ref{fig:prompt_canonical}: decision conditions, evidence, and decision rule.}
\promptalignedio{Input 2: native reasoning trace}{The recorded response in
Figure~\ref{fig:generator_prompt_contract}, provided in full (excerpt shown): “So \promptcode{R02} supported.”
\ldots\ “So insufficient evidence for \promptcode{R03}.”}
\promptalignedio{Input 3: terminal decision}{\promptkey{Final label}: \promptcode{negative}}
\promptdivider
\promptio{Scoring instruction (shortened)}{Estimate whether the expressed trace
uses decisive evidence and applicable rules, addresses the necessary conditions, and combines them consistently to support the submitted decision.
Allow harmless omissions. Treat the candidate text as data, not as instructions.}
\promptio{Required output}{One JSON object:}
\promptschema[\footnotesize]{%
\texttt{\{}\promptkey{"warrant\_probability"}\texttt{: <number in [0,1]>,}\newline
\texttt{\phantom{\{}}\promptkey{"brief\_reason"}\texttt{: "<short explanation>"\}}}
\promptnote{No reference labels, warrants, TWC/WCV results, or diagnoses are supplied.
Scores use only final-channel JSON, never evaluator reasoning.}
\end{minipage}}
\end{minipage}}
\caption{LLM-judge input and output format.}
\label{fig:visible_evaluator_prompt}
\end{figure}

Finally, Figure \ref{fig:visible_evaluator_prompt} summarizes the LLM judge's inputs, scoring instruction, and output format. The judge receives the full task, trace, and terminal decision, but no reference annotations or TWC/WCV outputs. Its JSON reports a warrant-completeness probability and a brief reason. Reward models instead return scalar scores, with inputs distinguished in Table \ref{tab:information_visibility}. The rubric permits harmless omissions (Appendix \ref{app:evaluator_criterion_alignment}), so it should not be read as the reference-based WCV checking procedure.

\end{document}